\documentclass[acmtog]{acmart}
\renewcommand\footnotetextcopyrightpermission[1]{} 
\acmSubmissionID{927}

\usepackage{booktabs} 

\usepackage[ruled]{algorithm2e} 

\SetAlFnt{\small}
\SetAlCapFnt{\small}
\SetAlCapNameFnt{\small}
\SetAlCapHSkip{0pt}

\usepackage{xcolor}
\usepackage{ifthen}
\usepackage{wrapfig}

\definecolor{Violet}{rgb}{.6, .35, .7}

\definecolor{JinboColor}{rgb}{0.15, 0.38, 0.68}
\definecolor{YixinColor}{rgb}{0.15, 0.68, 0.38}
\definecolor{AlanColor}{rgb}{0.68, 0.15, 0.38}
\definecolor{Red}{rgb}{1, 0, 0}
\definecolor{Pink}{rgb}{1, 0, 1}

\newcommand{\JY}[1]{{\leavevmode\color{JinboColor} JY: #1}}
\newcommand{\YH}[1]{{\leavevmode\color{YixinColor} YH: #1}}
\newcommand{\alan}[1]{{\leavevmode\color{AlanColor} alan: #1}}

\newcommand{\toref}[1]{\textcolor{Red}{[ref:#1]}}
\newcommand{\tocite}[1]{\textcolor{Pink}{[cite:#1]}}
\newcommand{\todo}[1]{\textcolor{Red}{TODO: #1}}

\newcommand{\nothing}[1]{}

\providecommand{\finalversion}{0} 
\ifthenelse{\equal{\finalversion}{1}}
{
	\renewcommand{\JY}[1]{}
	\renewcommand{\YH}[1]{}
        \renewcommand{\alan}[1]{}
	\renewcommand{\toref}[1]{}
	\renewcommand{\tocite}[1]{}
	\renewcommand{\todo}[1]{}
}
{}

\providecommand{\norevision}{1} 
\ifthenelse{\equal{\norevision}{1}}
{

}
{}

\usepackage{multirow}
\usepackage{amsmath}
\begin{document}
\title{Dragen3D: Multiview Geometry Consistent 3D Gaussian Generation with Drag-Based Control
}
\author{Jinbo Yan}
\affiliation{%
 \institution{Tencent}
 \country{China}}
\email{yanjbcool@gmail.com}
\author{Alan Zhao}
\affiliation{%
 \institution{Tencent}
 \country{China}
}
\email{alantzhao@tencent.com}
\author{Yixin Hu}
\affiliation{%
\institution{Tencent America}
\country{USA}}
\email{yixinhu.yh@gmail.com}


\begin{abstract}

Single-image 3D generation has emerged as a prominent research topic, playing a vital role in virtual reality, 3D modeling, and digital content creation. However, existing methods face challenges such as a lack of multi-view geometric consistency and limited controllability during the generation process, which significantly restrict their usability.
To tackle these challenges, we introduce \textsc{Dragen3D}, a novel approach that achieves geometrically consistent and controllable 3D generation leveraging 3D Gaussian Splatting (3DGS).
We introduce the Anchor-Gaussian Variational Autoencoder (Anchor-GS VAE), which encodes a point cloud and a single image into anchor latents and decode these latents into 3DGS, enabling efficient latent-space generation.
To enable multi-view geometry consistent and controllable generation, we propose a Seed-Point-Driven strategy: first generate sparse seed points as a coarse geometry representation, then map them to anchor latents via the Seed-Anchor Mapping Module. Geometric consistency is ensured by the easily learned sparse seed points, and users can intuitively drag the seed points to deform the final 3DGS geometry, with changes propagated through the anchor latents.
To the best of our knowledge, we are the first to achieve geometrically controllable 3D Gaussian generation and editing without relying on 2D diffusion priors, delivering comparable 3D generation quality to state-of-the-art methods.
\end{abstract}

%
%

%
%


\begin{teaserfigure}
  \centering
  \includegraphics[width=\textwidth]{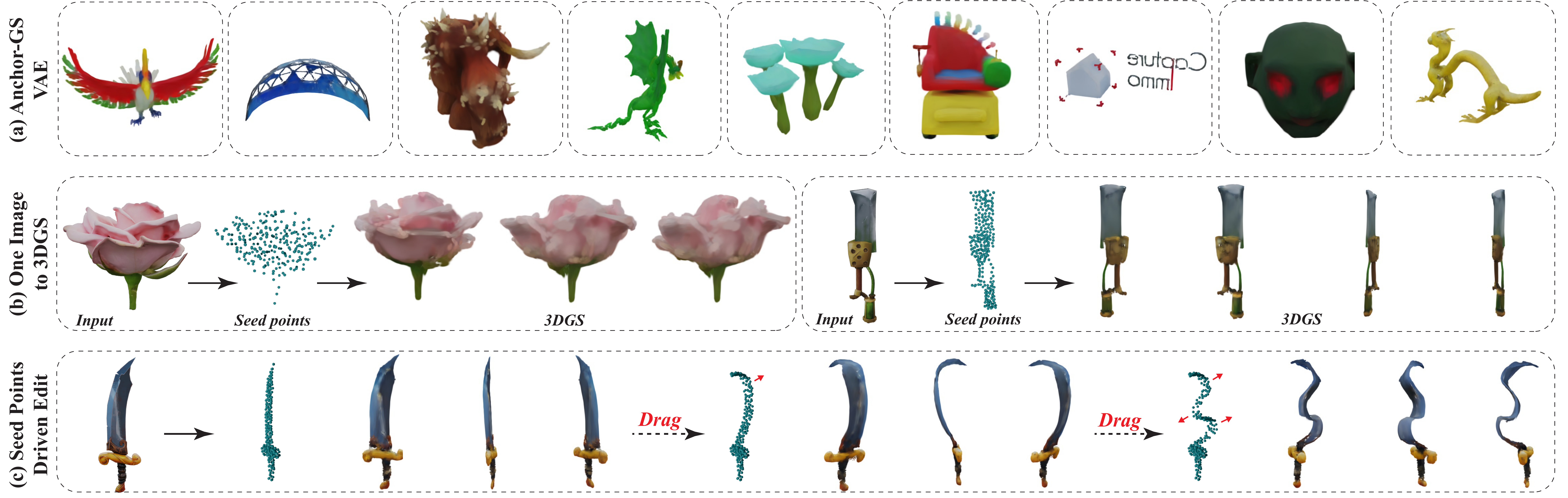} 
  \caption{We propose the Anchor-GS VAE and Seed-Point-Driven Generation strategy, which achieve high-quality 3DGS reconstructions (a) and multi-view geometry consistent single-image 3DGS generation (b). Additionally, we enable controllable 3DGS generation by allowing drag-based editing of the seed points (c).}
  \label{fig:teaser}
\end{teaserfigure}

\maketitle

\section{Introduction}

The field of 3D generation is highly popular at present and enjoys a wide range of applications in research and industry scenarios. However, compared to the traditional 3D modeling process where artists can directly interact and edit high-quality 3D models, achieving high geometric fidelity and direct editing within the 3D generation process is still an area awaiting in-depth study.

This challenge becomes even more pronounced in the context of 3D model generation from single-view images. For parts of the model not visible in the input image, the generated results may exhibit significant stylistic discrepancies from the visible regions, fail to achieve multi-view geometric consistency, or even appear unrealistic.
To align with the creative aspirations and modeling requirements of artists, some studies, as discussed in Sec.\ref{sec: relatd_edit}, have explored user control through input image modifications or predefined editing operations, these methods do not effectively address the aforementioned issues. 
To enhance the practical usability and quality of generated 3D models, we aim to develop a method that enables multi-view geometry consistent 3D generation, while allowing users to directly adjust and control the 3D shape during the generation process.

%
To this end, we propose an innovative approach, \textsc{Dragen3D}, utilizing sparse seed points for manipulating the object shape represented by 3D Gaussians (3DGS) and enhancing the multi-view geometry consistency within the 3D generation framework.
%
%
To accomplish this, we train a Variational Autoencoder (VAE) that encodes the complex 3D information of an object into a compact latent space and accurately decodes it back into the 3D domain, while also supporting subsequent 3D generation in the latent space.
%
Then, we introduce a module tasked with generating 3D seed points corresponding to the objects depicted in the input image. This ensures the geometric consistency of the seed points, thanks to the easy learning of their sparse distribution.
Furthermore, a mapping module is incorporated to associate the information of seed points with the latent space of the VAE. 


Our experiments show that \textsc{Dragen3D} produces multi-view geometry consistent 3D results as shown in Fig. \ref{fig:image-3d}. When the seed points undergo deformation, the corresponding latent codes are updated accordingly, enabling the generation of the final deformed 3D output upon decoding, as shown in Fig. \ref{fig:edit}

Our contributions can be summarized as follows:  
\begin{itemize}

    \item We propose the Anchor-GS VAE, which encodes 3D geometry and appearance into anchor latents and decodes them into 3DGS, making it easy to build while enabling efficient latent-space generation. 
    \item We introduce a Seed-Driven Strategy that generates sparse seed points from a single image for geometric consistency and maps them to anchor latents via the Seed-Anchor Mapping Module. 
    \item We design a Seed-Points-Driven Deformation module, enabling user-friendly geometric editing of 3DGS through drag operations on seed points. 
\end{itemize}
We will open-source the implementation of our method and the trained models.

\section{Related Work}

\subsection{Neural Rendering and Gaussian Splatting}
Radiance fields have become a popular research topic in 3D representation due to their powerful potential in 3D reconstruction and view synthesis. NeRF\cite{mildenhall2021nerf}, as a milestone work, made high-quality view synthesis possible. Its variants focus on improving rendering quality\cite{barron2021mip, barron2022mip,barron2023zip}, training and inference speed~\cite{muller2022instant,fridovich2022plenoxels,hedman2021snerg,SunSC22}, and generalization ability~\cite{wang2021ibrnet,yu2021pixelnerf,chen2021mvsnerf,johari2022geonerf}. Among them, 3D Gaussian Splatting (3DGS)~\cite{kerbl3Dgaussians} adopts a point-based radiance field, using 3D Gaussian primitives to represent scenes. Through anisotropic splatting and advanced rendering techniques, it enables high-quality reconstruction and real-time rendering. Some variants further enhance rendering quality and geometry~\cite{zhang2024rade,huang20242d, Yu2024GOF,lu2024scaffold,yu2024mip}, offering the ability to represent both high-quality geometry and textures, which provides solutions for various tasks and applications, including 3D generation.

\subsection{2D Diffusion Priors Based 3D Generation}
\label{sec:related-2d-diffusion}
Leveraging the high-quality generation capabilities of text-to-image diffusion models \cite{rombach2022high,saharia2022photorealistic,betker2023improving}, some multiview diffusion models \cite{liu2023zero,shi2023zero123++,shi2023mvdream,wang2023imagedream,li2023sweetdreamer,long2024wonder3d} enable view synthesis based on text/image and view conditions, facilitating 3D generation from 2D diffusion priors. Some methods optimize 3D representations from these 2D priors using an SDS-loss-based approach \cite{shi2023mvdream,liang2024luciddreamer,poole2022dreamfusion,wang2024prolificdreamer,tang2023dreamgaussian}or direct optimization~\cite{tang2025mvdiffusion++} from generated images. 
However, these methods are computationally expensive due to scene-by-scene optimization. Alternatively, other methods adopt a feed-forward~\cite{xu2024grm, xu2024instantmesh,li2023instant3d,liu2024one,tang2025lgm,chen2025lara} or denoising process~\cite{wang2025crm,liu2024one++,xu2023dmv3d} for 3D generation from 2D priors. For instance, LGM generates four views through a multiview diffusion model and then infers the corresponding 3DGS. These 2D-prior-based methods are constrained by inconsistencies in the multiview diffusion model, leading to misaligned 3D geometry and textures, and due to the stochastic nature of multiview image generation, they lack controlled generation.

\subsection{End-to-end 3D Generative Models}
\label{sec:related_3d}
Some methods \cite{tochilkin2024triposr,zou2024triplane,hong2023lrm} directly generate 3D representations from a single image without relying on 2D diffusion priors. For example, TriplaneGaussian~\cite{zou2024triplane} creates a point cloud from a single image, combines it with triplane fusion for texture, and produces the final 3DGS, achieving state-of-the-art single-image 3D results. Other approaches~\cite{zhang2024clay,zhang20233dshape2vecset,zhao2024michelangelo,gupta20233dgen,nichol2022point,muller2023diffrf} use 3D diffusion models, like 3DShape2VecSet~\cite{zhang20233dshape2vecset}, which encodes 3D information into a latent set and decodes it into a mesh, with diffusion models generating the latent set. Some approaches~\cite{zhang2024gaussiancube,zhou2024diffgs,he2025gvgen,xiang2024structured} also explore diffusion-based generation with Gaussian Splatting, such as GaussianCube~\cite{zhang2024gaussiancube}, which constructs structured Gaussian representations and uses a 3D U-Net-based diffusion model to generate Gaussians from noise. While these methods model 3D data distribution well, they lack user-friendly control for generation and editing. In contrast, our model leverages a diffusion model to learn 3D information distribution without needing a 3DGS dataset, offering controllable generation through 3D space manipulation.

\subsection{Editing in 3D Generative Models}  
\label{sec: relatd_edit}
To enable controllable 3D generation and editing, SketchDream~\cite{liu2024sketchdream} allows users to modify the sketch and achieve edits using SDS optimization for vivid results. However, its controllability is limited as user modifications are made in 2D space, which may not produce the desired effect for unselected viewpoints.  Interactive3D~\cite{dong2024interactive3d} directly edits 3DGS in 3D space using SDS optimization and predefined operations, converting the 3DGS representation into InstantNGP~\cite{muller2022instant} with further refine. MVDrag3D~\cite{chen2024mvdrag3d} projects 3D-space drag operations onto multiview images, using 2D diffusion editing capabilities, and infers the edited 3DGS through LGM~\cite{tang2025lgm}, followed by SDS refinement. These methods offer a more user-friendly experience. However, all of these methods rely on 2D generative priors, which may lead to geometric inconsistencies (as discussed in Sec. \ref{sec:related-2d-diffusion}), and require time-consuming optimization. In contrast, our method enables interactive manipulation of sparse seed points in 3D space, applying seed-point-driven deformation to modify the 3DGS without 2D priors or additional optimization, offering a more user-friendly editing experience.

\begin{figure*}[hbt]
  \centering
  \includegraphics[width=\textwidth]{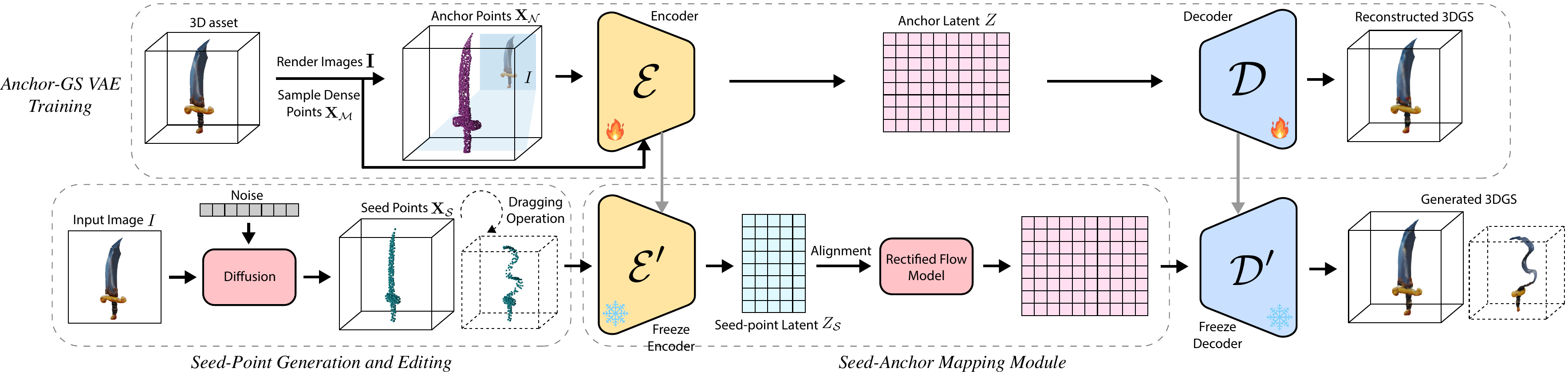} 
  \caption{Overview of the framework.}
  \label{fig:overview}
\end{figure*}

\section{Method}
\subsection{Overview}
Our method, \textsc{Drangen3D}, takes an image as input and generate a 3D object represented by 3D Guassians with multi-view geometric consistency, allowing user interaction of editing the geometry during the process. As illustrated in Fig.~\ref{fig:overview}, we first train an Anchor-Gaussian (Anchor-GS) VAE that encodes complex 3D information into a latent space and decodes it into 3DGS, enabling subsequent generation in the latent space (Sec.~\ref{sec:anchor-vae}).  
Then, we propose Seed-Point-Driven Controllable Generation module for 3D generation from a single image. This module starts with the generation of the rough initial geometry represented by a set of sparse surface points, named seed points, where we can apply the editing by deforming the seed points. After that, a mapping module is designed to map the (edited) seed point information to the latent space, which can be decode to 3DGS subsequently (Sec.~\ref{sec:seed-point-driven}). 





\subsection{Background}
\paragraph{Gaussian Splatting}
Gaussian splatting represents scenes as a collection of anisotropic 3D Gaussians. Each Gaussian primitive $\mathcal{G}_i$ is parameterized by a center $\mu \in \mathbb{R}^3$, opacity $\alpha \in \mathbb{R}$, color $c \in \mathbb{R}^{3(n+1)^2}$ which is represented by n-degree SH coefficients and 3D covariance matrix $\Sigma \in \mathbb{R}^{3 \times 3}$,which can be represented by scaling $s\in \mathbb{R}^3$ and rotation $r\in \mathbb{R}^4$.


During rendering, the 3D Gaussian is first projected onto 2D space. Given a view transformation matrix $W$, the 2D covariance matrix $\Sigma'$ can be computed as :
$\Sigma' = JW\Sigma W^T J^T$, where $J$ is the Jacobian of the affine approximation of the projective transformation. Subsequently, the Gaussians covering a pixel are sorted based on depth. The color of the pixel is obtained using point-based alpha blending rendering:
\begin{equation}
  c = \sum_{i=1}^n c_i \alpha_i \prod_{j=1}^{i-1}(1-\alpha_i)
  \label{3dgs_render}
\end{equation}

\paragraph{Rectified Flow Model}
The Rectified Flow Model \cite{liu2022flow, lipman2022flow} has the capability to establish a mapping between two distributions, \( \pi_0 \) and \( \pi_1 \), making it well-suited for our task of mapping seed point latents to anchor latents. Given \( x_0 \sim \pi_0 \) and the corresponding \( x_1 \sim \pi_1 \), we can obtain \( x(t) = (1 - t) x_0 + t x_1 \) at timestamp \( t \in [0,1]\) through linear interpolation. A vector field \( v_{\theta} \) parameterized by a neural network is used to drive the flow from the source distribution \( \pi_0 \) to the target distribution \( \pi_1 \) by minimizing the conditional flow matching objective:
\begin{equation}
    L(\theta) = E_{t,x_0,x_1,y}||v_{\theta}(x_t, t,y) - (x_1 - x_0)||
    \label{eq:flow matching}
\end{equation}
Here, $v_{\theta}(x_t, t, y)$ is the predicted flow at time $t$ for a given point $x_t$, $y$ refers to the image  condition that guides the flow matching.
\section{Anchor-Based 3DGS VAE}
\label{sec:anchor-vae}


We adopt an anchor-based approach to obtain 3D Gaussians, where the ``anchor'' refers to anchor points that are surface points capturing the main geometry of the object. 
%
We design and train an Anchor-Gaussian VAE that utilize 
%
Geometry-Texture Encoder $\mathcal{E}$ to encode 
geometry and appearance information
of a 3D object 
into a set of fixed length latents, called anchor latents $Z$ (Sec. \ref{sec:enc}). Subsequently, the Decoder $\mathcal{D}$  decodes these anchor latents into Gaussian primitives in a coarse-to-fine manner (Sec. \ref{sec:dec}). The encoder and decoder are trained together in an end-to-end manner, with the loss function (Sec. \ref{sec:loss}).



\subsection{Geometry-Texture Encoder}

The Geometry-Texture Encoder encodes the anchor points, the surface point cloud, and a set of rendered images of an object into a latent space.
We obtain the anchor points $\mathbf{X}_\mathcal{N}$ of an object by sampling from the surface point cloud $\mathbf{X}_\mathcal{M} \in \mathbb{R}^{M\times 3}$ of a 3D object using Farthest Point Sampling (FPS) method, which is similar to \cite{zhang20233dshape2vecset, zhang20223dilg}.
Here \( \mathcal{N} \subsetneqq \mathcal{M} \), represents the index set of point clouds, with default settings of $|\mathcal{N}|=2048$ and $ |\mathcal{M}|=4096$, and \( \mathbf{X}_\mathcal{N} \in \mathbb{R}^{N \times 3} \) denotes the sampled anchor points. 

To encode the appearance information, we then project these anchor points onto the image feature plane $P_I \in \mathbb{R}^{H\times W \times C}$, which is encoded from the rendered image \(I\) of a known viewpoint:
$
    \forall i \in \mathcal{N},\; \; \;  f_{i}=\Psi( \Pi_I(x_i), P_I)
$
, where \( \Pi_I(x_i) \) represents the projection of \( x_i\) onto the image plane of \( I\) using the camera parameters of \( I\), and \( \Psi \) denotes bilinear interpolation. 
%
%
The \( f_i \) and positional encoding of \( x_i \)  represents the  texture information and geometric of the \( i \)-th anchor. 

To allow each anchor to capture more global information, we then input these features into two layers of Transformer blocks, which utilize point clouds $\mathbf{X}_\mathcal{M}$ and image tokens extracted from the input image \( I \) to perform cross-attention:
\begin{equation}
\begin{aligned}
   & Z^{'} = \mathbf{Transformer}_1(\{(\mathbf{PE}(x_i);f_i) \}_{i \in \mathcal{N}}| \{\mathbf{PE}(x_i)\}_{i \in \mathcal{M}}) \\
& Z = \mathbf{Transformer}_2(Z^{'}| F_I) 
\end{aligned}
\label{eq: enc}
\end{equation}
where $Z$ represents the anchor latents obtained through encoding, \textbf{PE} represents the positional encoding, and (;) denotes concatenation along the channel dimension. $F_I \in \mathbb{R}^{N\times C}$ refers to the image feature tokens extracted by the Image Encoder from the input image $I$, where we use DINOv2\cite{oquab2023dinov2} for the feature extraction.
And $\mathbf{Transformer}(|)$ denotes a Transformer block with cross-attention.
All these encoding processes can be collectively represented by $\mathcal{E}$:
\begin{equation}
    Z = \mathcal{E}(\mathbf{X}_\mathcal{N} \mid \mathbf{X}_\mathcal{M} , I)
\end{equation}
After passing through the encoder $\mathcal{E}$, the anchor feature 
$ Z$ simultaneously encodes both geometric and texture information.
\label{sec:enc}

\subsection{Decoder}
\label{sec:dec}
In the Decoder, we adopt a coarse-to-fine approach to progressively obtain the Gaussian primitives, which enables higher-quality and more complete geometry. In the Encoder $\mathcal{E}$, both geometry and texture information are consolidated into a set of anchor latents \( Z \), which is first decoded into a coarse geometry and then refined to recover more detailed geometry and corresponding textures.

Specifically, we apply Transformer with self-attention to \( Z \):
\begin{equation}
    Z^L = \textbf{Transformer}(Z)
\end{equation}
Here, the Transformer block consists of \( L \) layers, and \( \{ Z^j \}_{j=1..L} \) represents the output at $j$-th layer of the Transformer with $Z^L$ being the final output.  We select the output from the \( k \)-th layer ($k=2$ and $L=8$ in default) as \( Z^{\text{coarse}} \), and the output from the last layer as \( Z^{\text{fine}} \). We first pass \( Z^{\text{coarse}} \) through a linear layer to reconstruct the anchors' spatial positions:
\[
\hat{\mathbf{X}}_\mathcal{N}
 = \textbf{Linear}(Z^{\text{coarse}})
\]

The symbol \( \hat{\mathbf{X}}_\mathcal{N} \in \mathbb{R}^{N \times 3} \) represents the reconstructed positions of anchor points, which approximates the coarse geometry. Then, we assign \( m \) ($m=8$ in default) Gaussian points to each anchor point. The positions of these Gaussian points are determined based on the anchor points' positions and a set of offsets derived from \( Z^{\text{fine}} \). For the \( i \)-th anchor point, we have:
\begin{equation}
\begin{aligned}
     \{ O_i^1, \dots, O_i^m \} &= \textbf{Linear}(z_{i}^{\text{fine}}) \\
     \{ \mu_i^1, \dots, \mu_i^m \} &=  \hat{x}_i+ \{ O_i^1, \dots, O_i^m \}
\end{aligned}
\end{equation}
where \( z_{i}^{\text{fine}} \) is the fine feature of $i$-th anchor and \( \hat{x}_i \) represents the coarse position decoded for the \( i \)-th anchor point. Here, \( \{ O_i^j \}_{j=1..m} \) denotes the offsets of the \( j \)-th Gaussian point relative to the anchor position, and \( \{ \mu_i^j \}_{j=1..m} \) represents the final positions of the Gaussian points. This way, we obtain a set of Gaussian point positions with dimensions \(  \mathbb{R}^{N' \times 3} \), where \( N' = N \times m \), representing the final fine-grained geometry.

For each Gaussian point, we can assign its other attributes by interpolating from its $k$ ($k=8$ in default) nearest anchors in the neighborhood:
\begin{equation}
    z_i = \frac{\sum_{k \in \mathcal{N}(\mu_i)} e^{-d_k}z_{k}^{fine}}{\sum_{k \in \mathcal{N}(\mu_i)}e^{-d_k}}
    \label{interpolate}
\end{equation}
where $\mathcal{N}(\mu_i)$ represents the set of neighboring anchor points of Gaussian point position $\mu_i$, and $d_k$ represents the Euclidean distance from $\mu_i$ to the reconstructed position of $z_k^{fine}$.
Then we can use a linear layer to decode the attributes color $c_i$, opacity $o_i$, scale $scale_i$, and rotation $rot_i$ of a Gaussian primitive $z_i$: $\{ c_i, o_i, scale_i,rot_i\} = \mathbf{Linear}(z_i)$.


\subsection{Loss Function}
\label{sec:loss}
Thanks to our efficient anchor-based representation design, the training of our VAE does not require pre-constructing a large-scale 3DGS dataset. Instead, we supervise the entire network using the rendering loss between the predicted rendered images and the ground truth images:
\begin{equation}
    \mathcal{L}_{\text{rgb}} = \mathcal{L}_{\text{MSE}} + \lambda_s \mathcal{L}_{\text{SSIM}} + \lambda_l \mathcal{L}_{\text{lpips}}
\end{equation}
where $\mathcal{L}_{\text{MSE}}$represents the pixel-wise Mean Squared Error (MSE) loss, $\mathcal{L}_{\text{SSIM}}$ represents the Structural Similarity Index (SSIM) loss and $\mathcal{L}_{\text{lpips}}$ represents the perceptual loss.

In addition, to obtain better geometry, we apply 3D point cloud supervision to both the reconstructed anchor point positions and the Gaussian point positions, comparing them with ground-truth points sampled from the 3D assets:
\begin{equation}
    \mathcal{L}_{\text{points}} = \lambda_c \mathcal{L}_{\text{cd}} + \lambda_e \mathcal{L}_{\text{emd}}
\end{equation}

Here, \( \mathcal{L}_{\text{cd}} \) denotes the Chamfer Distance (CD), and \( \mathcal{L}_{\text{emd}} \) represents the Earth Mover's Distance (EMD).

Finally, incorporating KL divergence regularization on the anchor latents produced by the encoder, the total loss function is defined as:
\begin{equation}
    \mathcal{L} = \mathcal{L}_{\text{rgb}} + \mathcal{L}_{\text{points}} + \lambda_{\text{kl}} \mathcal{L}_{\text{KL}}
\end{equation}

\section{Seed-Point-Driven Anchor Latent Generation and Editing}
\label{sec:seed-point-driven}

We adopt a Seed-Point-Driven generation approach to progressively obtain the anchor latents $Z$. First, we generate a sparse set of seed points \( \mathbf{X}_\mathcal{S} \in \mathbb{R}^{S \times 3} \) , which can be viewed as a rough representation of the geometry(Sec.~\ref{sec:seed-gen}). And then, through the Seed-Anchor Mapping module, we transform the sparse distribution of seed points into a dense distribution of anchor latents (Sec.~\ref{sec:seed-mapping}). The Seed-Point-Driven strategy enables interactive geometric control of the generated 3DGS by simply \emph{dragging} the seed points (Sec.~\ref{sec:seed-drag}).


This approach has the following advantages: (1)\textbf{Geometrically Consistent Generation}:We first learn a sparse set of seed points \( \mathbf{X}_\mathcal{S} \) (\( S = 256 \)), which ensures geometrically consistent 3D results due to the sparse nature of seed points and the ease of learning their distribution. (2) \textbf{Support for Geometric Editing}: By constructing the Seed-Anchor Mapping Module, we map seed points to their corresponding anchor latents. This decoupled design naturally supports geometric editing—modifying the seed points results in different anchor latents, enabling deformation of the 3DGS. 

\subsection{Seed Points Generation Module}

\label{sec:seed-gen}

\begin{figure}
  \includegraphics[width=\columnwidth]{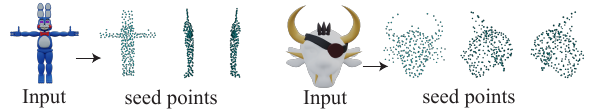}
  \caption{
Generation of Seed Points with Multiview Geometry Consistency
  }
  \label{fig:seed_gen}
\end{figure}

Our goal is to generate a sparse set of seed points $\mathbf{X}_\mathcal{S}$ as a rough representation of the geometry from a single image input. To achieve this, we employ a diffusion model conditioned on the image to learn the distribution of $\mathbf{X}_\mathcal{S}$. Given the sparse nature of the seed points \(\mathbf{X}_\mathcal{S}\), where \(|\mathcal{S}| = 256\) in our settings, their distribution is relatively simple to learn directly, without the need for projection into a latent space. The results can be seen in Fig. \ref{fig:seed_gen}. Specifically, we utilize the Rectified Flow model to map Gaussian noise to the seed point distribution \(\pi_S\), treating the noise \(\epsilon\) as \(x_0\) and the data sample \(x_S\) as \(x_1\).


\subsection{Seed-Anchor Mapping Module}
\label{sec:seed-mapping}
\begin{figure}
  \includegraphics[width=\columnwidth]{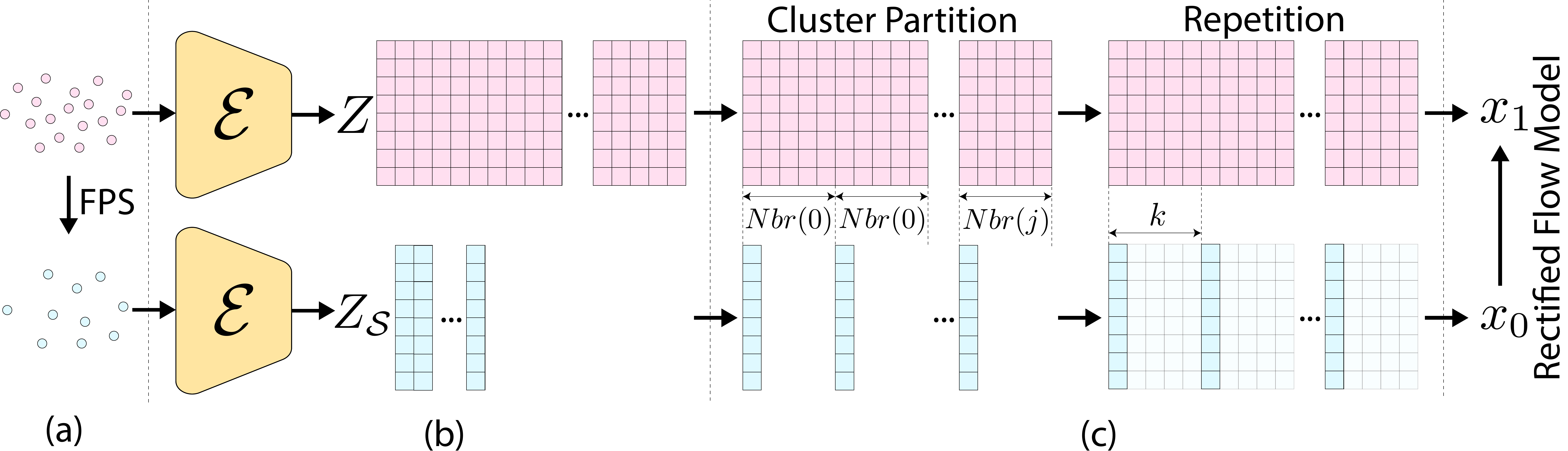}
  \caption{
  Seed-Anchor Mapping Module: 
(a) We use FPS to establish a correspondence between \( Z \) and \( \mathbf{X}_\mathcal{S} \).  
(b) Dimension Alignment: Encoding the seed points \( \mathbf{X}_\mathcal{S} \) to obtain \( Z_\mathcal{S} \), ensuring dimensional alignment with \( Z \).  
(c) Token Alignment: Each token in the seed latent is treated as a center to partition the tokens of \( Z \) into \( |\mathcal{S}| \) clusters. A repeat operation is then applied to the seed latents, achieving semantic and token count alignment between \( Z_\mathcal{S} \) and \( Z \).
  }
  \label{fig:token_align}
\end{figure}

To use an input image $I$ and a set of (deformed) seed points to control the generation of 3DGS, we need to derive the corresponding anchor latents \( Z \).
We model this task as a flow matching problem between two distributions and aim to solve it using the Rectified Flow Model,  as shown in Fig. \ref{fig:token_align}.

First, we need to establish a one-to-one correspondence between known samples from these two distributions.  Specifically, for each anchor point set \( \mathbf{X}_\mathcal{N} \), we apply Furthest Point Sampling (FPS) to downsample the anchor points to obtain the seed points \( \mathbf{X}_\mathcal{S} \). That ensures for each \( Z \), we can find a corresponding \( \mathbf{X}_\mathcal{S} \).

\paragraph{Dimension Alignment}
To construct the Rectified Flow model, the starting and targets must share the same dimensionality. Thus, we encode the seed points to align with the dimension of \( Z \) by passing \( \mathbf{X}_\mathcal{S} \) through the freeze encoder \( \mathcal{E} \) of Anchor-GS VAE:
\begin{equation}
  Z_\mathcal{S} = \mathcal{E}(\mathbf{X}_\mathcal{S}| \mathbf{X}_\mathcal{S}, I)  
  \label{eq:encode_seed_latents}
\end{equation}
Here, \( Z_\mathcal{S} \) represents the encoded latents of the seed points. This allows us to simplify the problem into a mapping from \( Z_\mathcal{S} \) to corresponding \( Z\). Using the same pretrained encoder with \( Z \), the \( Z_\mathcal{S} \) distribution becomes better aligned with the target anchor latents \( Z \), providing valuable information about the alignment between the points and the image.

\paragraph{Token Alignment.}
To establish flow matching between \( Z_\mathcal{S} \)and \( Z\), we need to ensure that both samples contain the same number of tokens, and each token in the two samples corresponds semantically. Unlike \cite{fischer2023boosting}, which performs 2D grid-based upsampling on images, we cannot simply upsample seed points to match the target size while maintaining semantic correspondence between the points, as our latents are unordered.

To address this, we propose a cluster-based token alignment strategy. Each token in the latents retains the geometric information of the encoded points, allowing us to partition the latents into clusters based on their spatial positions. Specifically, for each token in the seed latents, we identify its neighborhood in the anchor latents using:
\begin{equation}
    \forall i \in \mathcal{S}, \quad \text{KNN}(x_i) = \{x_j\}_{j \in \text{Nbr}(i)}
\end{equation}
Here, \( x_j,x_i\) represent the encoded position of $z_j \in Z$ and $z_i \in Z_\mathcal{S}$, respectively, while $\text{Nbr}(i)$ denotes the index set of the $k$-nearest neighbors around \( z_i\).
This partitions \( Z\) partition into $|\mathcal{S}|$ clusters, where the tokens in  each cluster $\text{Nbr}(i)$ are semantically similar to $z_i$  , leveraging their spatial proximity. 
After establishing the semantic similarity between each token in $Z_\mathcal{S}$ and the corresponding cluster of tokens in $Z$, we simply repeat tokens in $Z_\mathcal{S}$ to ensure numerical equivalence. At this point, the alignment results between \( Z \) and \( Z_\mathcal{S} \) can serve as the start point $x_0 $and target $x_1$ for the Rectified Flow model, with the same number of tokens and semantic correspondence.

\paragraph{Model Architecture and Details.}
We implement the model using Transformer blocks, with image conditions serving as the key and value in the cross-attention, and inject timestamps via the adaptive layer norm (adaLN) block as described in \cite{peebles2023scalable}. With token alignment, the input tokens \( x_t \) are clustered and exhibit spatial similarity within each cluster. Therefore, we can downsample and then upsample within each cluster to reduce computational complexity, skipping connections to transfer detailed information.
Similar to \cite{deng2024detailgen3d,fischer2023boosting}, we apply noise augmentation to the start point \( x_0 \), which enhances the stability of our training process. Additionally, since the seed points used during inference are generated and subject to various edits, this noise augmentation helps our model generalize to a wider variety of seed points. We apply a cosine schedule for noise augmentation at 150 timesteps during both training and evaluation.

\subsection{Seed-Points-Driven Deformation}
\label{sec:seed-drag}

Thanks to our Seed-Anchor Mapping module, the mapping  begins with seed points—a sparse set of points that guide and control the overall 3D geometry generation. By adjusting the positions of these seed points, we can intuitively generate various desired geometries, making the editing process flexible and precise. The discrete nature of point clouds enables effective application of drag-style editing. Additionally, mature 3D tools like Blender\cite{blender} already support such operations on point clouds in 3D space, providing an intuitive and user-friendly editing experience.

Specifically, for the initial seed points \( X_\mathcal{S} \) and their corresponding \( Z_\mathcal{S} \), we apply drag-style editing to the seed points, resulting in \( X_{\hat{S}} \). We then encode \( X_{\hat{S}} \) to obtain \( \hat{Z}_\mathcal{S} \), using Eq. \ref{eq:encode_seed_latents}. During encoding, we continue to use the projected features obtained from the projection of \( X_\mathcal{S} \) onto the input image to preserve the correspondence between geometry and texture.

To preserve the consistency of these unedited regions, we introduce a mask to ensure their invariance:
\begin{equation}
    Z^{*}_\mathcal{S} = mask \odot Z_\mathcal{S} + (1 - mask)\odot \hat{Z}_\mathcal{S}
\end{equation}
In this equation, the mask is a Boolean vector indicating whether a point remains unchanged. 
With this, \( Z^{*}_\mathcal{S} \) serves as the new seed latents. By applying the same alignment operation and Seed-Anchor Mapping module, we derive the corresponding anchor latents from the dragged seed points, which are then decoded into the deformed 3DGS. This process ensures that the dragged points remain aligned with their original texture while maintaining consistency in the unedited regions.

\section{Experiments}
\subsection{Implementation Details}
\label{sec:implemenet}
\paragraph{Datasets} We use the Multiview Rendering Dataset \cite{qiu2023richdreamer,zuo2024sparse3d} based on Objaverse \cite{objaverse} for training. The dataset includes 260K objects, with 38 views rendered for each object, with a resolution of $512\times512$. To obtain the surface point clouds, we transform the 3D models according to the rendering settings, filter out those that are not aligned with the rendered images, and use Poisson sampling method\cite{poisson} to sample the surface.  We randomly split the final processed data into training and testing sets, with the training dataset consisting of 200K objects.
We conduct our in-domain evaluation using the test set from Objaverse, which includes 2,000 objects. To assess our model's cross-domain capabilities, we evaluate it on the Google Scanned Objects (GSO) \cite{downs2022google}dataset, which contains 1,030 real-world scanned 3D models, with 32 views rendered for each model on a spherical surface.

\paragraph{Network}
In our implementation, the anchor latents have a fixed length of 2048 and a dimension of 8. The model dimension in our transformer blocks is 512, with each transformer block comprising two attention layers and a feed-forward layer, following the design in \cite{zou2024triplane}. The Anchor-GS VAE consists of two transformer blocks in the encoder and eight transformer blocks in the decoder. 
The Seed-Anchor Mapping Module is implemented using 24 transformer blocks, with 4 blocks for downsampling and 4 blocks for upsampling. Similarly, the Seed Points Generation Module is implemented with 24 transformer blocks. Leveraging the sparsity of seed points, we directly learn their distribution without requiring a VAE. The image conditioning in our model is extracted using DINOv2\cite{oquab2023dinov2}.

\paragraph{Training Details}
For training the Anchor-GS VAE, we randomly select 8 views per object, using one view as the input and all 8 views as ground truth images for supervision. The loss weights are set as \(\lambda_s = 1\), \(\lambda_l = 1\), \(\lambda_c = 1\), \(\lambda_e = 1\), and \(\lambda_{KL} = 0.001\). We train the Anchor-GS VAE on a subset of our collected dataset containing approximately 40K objects, using a batch size of 128 on 8 A100 GPUs (40GB) for 24K steps. The Seed-Anchor Mapping Module is trained on the full dataset with a batch size of 1280 on 64 V100 GPUs(32GB) for 20K steps. The Seed Points Generation Module is trained on 48 V100 GPUs (32GB) for 54K steps.  We use the AdamW optimizer with an initial learning rate of \(4 \times 10^{-4}\), which is gradually reduced to zero using cosine annealing during training. The sampling steps for both the Seed-Anchor Mapping Module and the Seed Points Generation Module are set to 50 during inference.

\paragraph{Baseline}
We compared our method with previous SOTA 3DGS generation models in the single-image input setting. LGM and LaRA rely on 2D multi-view diffusion priors to obtain multi-view images, which are then used to generate the output 3DGS in a feed-forward manner, as described in ~\ref{sec:related-2d-diffusion}. TriplaneGS~\cite{zou2024triplane} does not require a 2D diffusion prior, directly generating 3DGS from a single input image, as outlined in ~\ref{sec:related_3d}. Both of them achieving SOTA performance. For each compared method, we use the official models and provided weights and ensure careful alignment of the camera parameters.


\subsection{Results of VAE Reconstruction}
In Fig. \ref{fig:vae}, we present the results of our Anchor-GS VAE. Given point clouds and a single image, our Anchor-GS VAE achieves high-quality reconstructions with detailed geometry and textures.

\subsection{Results of 3D Generation }
\label{sec:comparison}
\paragraph{Metrics} 
Following previous works \cite{zou2024triplane, chen2025lara}, we use peak signal-to-noise ratio (PSNR), perceptual quality measure LPIPS, and structural similarity index (SSIM) as evaluation metrics to assess different aspects of image similarity between the predicted and ground truth. Additionally, we report the time required to infer a single 3DGS. We use a single image as input and evaluate the 3D generation quality using all available views as testing views to compare our method with others, all renderings are performed at a resolution of 512.

Tab. \ref{tab:quantitative comparison} presents the quantitative evaluation results of our method compared to previous SOTA methods on the Objaverse and GSO datasets, along with qualitative results shown in Fig. \ref{fig:image-3d}. The multi-view diffusion model used in LGM tend to produce more diverse but uncontrollable results, and lacks precise camera pose control. As a result, it fails in our dense viewpoints evaluation, achieving PSNR scores of 12.76 and 13.81 on the Objaverse and GSO test sets, respectively.

As shown in Tab. \ref{tab:quantitative comparison}, LGM and LaRa, influenced by the multi-view inconsistency of 2D diffusion models, achieve relatively lower scores in our dense viewpoint evaluation. In contrast, our method achieves the best results across both datasets, with only a slight overhead in inference time.

Fig. \ref{fig:image-3d} presents the first six rows from the Objaverse dataset and the last three rows from the GSO dataset. All methods are compared using the same camera viewpoints. For the Objaverse dataset, the rendering viewpoints are the left and rear views relative to the input viewpoint, while for the GSO dataset, the views are selected to showcase the object as completely as possible. Compared to methods using 2D diffusion priors, such as LGM and LaRa, our method demonstrates better multi-view geometric consistency, while the former tends to generate artifacts or unrealistic results in our displayed views. Compared to TGS, our method learns the 3D object distribution more effectively, resulting in more geometrically consistent multi-view results, such as the sharp feature in the left view in the first knife case.



\subsection{Editing Results Based on Drag}
As shown in Fig. \ref{fig:edit}, our method enables Seed-Points-Driven Deformation. Starting with generated seed points from the input image, the sparse nature of the seed points allows for easy editing using 3D tools (e.g., Blender\cite{blender}) with a few drag operations. The edited 3DGS can then be obtained within 2 seconds.
\subsection{Ablation Study}
\begin{figure}
  \includegraphics[width=\linewidth]{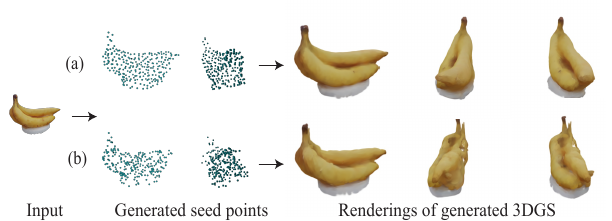}
  \caption{Ablation study about different seed points geneartion methods: (a) using our method, (b) using Transformer.}
  \label{fig:ablation-seed-gen}
\end{figure}

\paragraph{Seed Points Generation}
We employ a Rectified Flow model to generate seed points conditioned on a single input image. Owing to the sparsity of the seed points, the flow model is easier to train and effectively learns the distribution of the seed points. However, we also explored an alternative implementation using a transformer-based feed-forward approach, where point clouds are generated from learnable embeddings in a single pass, as in \cite{zou2024triplane}. As demonstrated in Fig. \ref{fig:ablation-seed-gen}, the feed-forward approach struggles to capture the true distribution of seed points and fails to produce satisfactory results in regions not visible in the input image.

\begin{figure}
  \includegraphics[width=\linewidth]{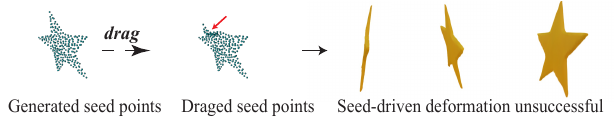}
  \caption{Without Dimension Alignment, seed-points-driven deformation fails}
  \label{fig:ablation-seed-enc}
\end{figure}

\paragraph{Dimension Alignment}
To match the dimension of the starting and target points in the Seed-Anchor Mapping Module, we encode the seed points using the Anchor-GS VAE encoder (Eq. \ref{eq:encode_seed_latents}). This process brings their distributions closer, reducing learning difficulty and reliance on image conditions. 
To validate this method, we conducted experiments by replacing the encoding approach with positional encoding .  When using positional encoding, the Seed-Anchor Mapping overly relied on the image condition, neglecting the contribution of the seed points and failing to enable seed-driven 3DGS deformation, as shown in Fig. \ref{fig:ablation-seed-enc}. 


\paragraph{Token Alignment}
We ensure token alignment in Flow Matching by organizing tokens around seed points, followed by  cluster-based partition and repetition. To evaluate its effectiveness, we conducted two ablation experiments, as shown in Tab. \ref{tab:ablation-tokenalign}. In the \textit{No-cluster+No-repetition} setting, we omitted the clustering step, aligning only the corresponding seed and anchor latents while filling unmatched portions with noise. This also prevented cluster-based downsampling in the Flow Model, resulting in doubled memory consumption. In the \textit{No-cluster} setting, we repeated the seed latents to match the number of anchor latents but left them unordered, leading to disorganized token matching. As shown in Tab. \ref{tab:ablation-tokenalign}, on a 40K subset with the same number of training steps, the absence of token alignment significantly degraded Flow Matching performance, resulting in inaccurate correspondences.
\begin{table}%
\caption{ Quantitative evaluation of our method compared to previous work. $\dagger$ achieves relatively lower PSNR values in the evaluation, so we display the results in Sec. \ref{sec:comparison}.}
\label{tab:quantitative comparison}
\resizebox{0.5\textwidth}{!}{
\begin{tabular}{llllllll}
  \toprule
  \multirow{2}{*}{Method}  & \multicolumn{3}{c}{Objaverse\cite{objaverse}}   & \multicolumn{3}{c}{GSO\cite{downs2022google}}& \multirow{2}{*}{Time(s)}\\
\cmidrule(r){2-4}  \cmidrule(r){5-7} 
   & PSNR$\uparrow$& SSIM$\uparrow$& LPIPS$\downarrow$ & PSNR$\uparrow$& SSIM$\uparrow$& LPIPS $\downarrow$
   \\ \midrule
  LGM$\dagger$\cite{tang2025lgm}     & -&0.836&0.211&-&0.833&0.21&4.82\\
  LaRa\cite{chen2025lara}  & 16.57&0.860&0.174&15.98&9.869&0.162&9.50\\
  TriplaneGS\cite{zou2024triplane}  &18.80 &0.883&0.143&19.84&0.900&0.120&0.70\\
  Ours &20.92&0.896&0.120&20.52&0.904&0.1122&4.71\\
  \bottomrule
\end{tabular}
}
\end{table}%

\begin{table}%
\caption{Ablation about token alignment}
\label{tab:ablation-tokenalign}
\begin{minipage}{\columnwidth}
\begin{center}
\begin{tabular}{llll}
  \toprule
   & PSNR$\uparrow$& SSIM$\uparrow$& LPIPS$\downarrow$ 
   \\ \midrule
  No-cluster+No-repetition  & 18.84&0.877&0.141\\
  No-cluster     & 19.20 &0.876&0.142\\
  ours-full  &19.94 &0.881&0.134\\
  \bottomrule
\end{tabular}
\end{center}
\bigskip\centering

\end{minipage}
\end{table}%

\section{Conclusions}

In this paper, we present \textsc{Dragen3D}, a framework for multi-view geometry consistent single-image 3DGS generation with drag-based editing.  
We propose the Anchor-GS VAE, which encodes 3D geometry and texture into anchor latents and decodes them into 3DGS. Combining seed-point generation from single image, user-interacted seed-point editing, and seed-to-anchor-latent mapping, we are able to generate and control the final output 3DGS. Evaluations across multiple datasets demonstrate that \textsc{Dragen3D} achieves state-of-the-art 3DGS quality from single images.
However, our method has room for improvement. First, incorporating 3D appearance editing based on prompts could be an interesting direction to explore, especially when integrated with existing multimodal large models. 
Additionally, the quality and quantity of training data limit our model's capabilities, which can be further improved with more computational resources.



\bibliographystyle{ACM-Reference-Format}
\bibliography{99-bib, my}


\begin{thebibliography}{70}


\ifx \showCODEN    \undefined \def \showCODEN     #1{\unskip}     \fi
\ifx \showDOI      \undefined \def \showDOI       #1{#1}\fi
\ifx \showISBNx    \undefined \def \showISBNx     #1{\unskip}     \fi
\ifx \showISBNxiii \undefined \def \showISBNxiii  #1{\unskip}     \fi
\ifx \showISSN     \undefined \def \showISSN      #1{\unskip}     \fi
\ifx \showLCCN     \undefined \def \showLCCN      #1{\unskip}     \fi
\ifx \shownote     \undefined \def \shownote      #1{#1}          \fi
\ifx \showarticletitle \undefined \def \showarticletitle #1{#1}   \fi
\ifx \showURL      \undefined \def \showURL       {\relax}        \fi
\providecommand\bibfield[2]{#2}
\providecommand\bibinfo[2]{#2}
\providecommand\natexlab[1]{#1}
\providecommand\showeprint[2][]{arXiv:#2}

\bibitem[Barron et~al\mbox{.}(2021)]%
        {barron2021mip}
\bibfield{author}{\bibinfo{person}{Jonathan~T Barron}, \bibinfo{person}{Ben Mildenhall}, \bibinfo{person}{Matthew Tancik}, \bibinfo{person}{Peter Hedman}, \bibinfo{person}{Ricardo Martin-Brualla}, {and} \bibinfo{person}{Pratul~P Srinivasan}.} \bibinfo{year}{2021}\natexlab{}.
\newblock \showarticletitle{Mip-nerf: A multiscale representation for anti-aliasing neural radiance fields}. In \bibinfo{booktitle}{\emph{Proceedings of the IEEE/CVF international conference on computer vision}}. \bibinfo{pages}{5855--5864}.
\newblock


\bibitem[Barron et~al\mbox{.}(2022)]%
        {barron2022mip}
\bibfield{author}{\bibinfo{person}{Jonathan~T Barron}, \bibinfo{person}{Ben Mildenhall}, \bibinfo{person}{Dor Verbin}, \bibinfo{person}{Pratul~P Srinivasan}, {and} \bibinfo{person}{Peter Hedman}.} \bibinfo{year}{2022}\natexlab{}.
\newblock \showarticletitle{Mip-nerf 360: Unbounded anti-aliased neural radiance fields}. In \bibinfo{booktitle}{\emph{Proceedings of the IEEE/CVF conference on computer vision and pattern recognition}}. \bibinfo{pages}{5470--5479}.
\newblock


\bibitem[Barron et~al\mbox{.}(2023)]%
        {barron2023zip}
\bibfield{author}{\bibinfo{person}{Jonathan~T Barron}, \bibinfo{person}{Ben Mildenhall}, \bibinfo{person}{Dor Verbin}, \bibinfo{person}{Pratul~P Srinivasan}, {and} \bibinfo{person}{Peter Hedman}.} \bibinfo{year}{2023}\natexlab{}.
\newblock \showarticletitle{Zip-nerf: Anti-aliased grid-based neural radiance fields}. In \bibinfo{booktitle}{\emph{Proceedings of the IEEE/CVF International Conference on Computer Vision}}. \bibinfo{pages}{19697--19705}.
\newblock


\bibitem[Betker et~al\mbox{.}(2023)]%
        {betker2023improving}
\bibfield{author}{\bibinfo{person}{James Betker}, \bibinfo{person}{Gabriel Goh}, \bibinfo{person}{Li Jing}, \bibinfo{person}{Tim Brooks}, \bibinfo{person}{Jianfeng Wang}, \bibinfo{person}{Linjie Li}, \bibinfo{person}{Long Ouyang}, \bibinfo{person}{Juntang Zhuang}, \bibinfo{person}{Joyce Lee}, \bibinfo{person}{Yufei Guo}, {et~al\mbox{.}}} \bibinfo{year}{2023}\natexlab{}.
\newblock \showarticletitle{Improving image generation with better captions}.
\newblock \bibinfo{journal}{\emph{Computer Science. https://cdn. openai. com/papers/dall-e-3. pdf}} \bibinfo{volume}{2}, \bibinfo{number}{3} (\bibinfo{year}{2023}), \bibinfo{pages}{8}.
\newblock


\bibitem[Chen et~al\mbox{.}(2025)]%
        {chen2025lara}
\bibfield{author}{\bibinfo{person}{Anpei Chen}, \bibinfo{person}{Haofei Xu}, \bibinfo{person}{Stefano Esposito}, \bibinfo{person}{Siyu Tang}, {and} \bibinfo{person}{Andreas Geiger}.} \bibinfo{year}{2025}\natexlab{}.
\newblock \showarticletitle{Lara: Efficient large-baseline radiance fields}. In \bibinfo{booktitle}{\emph{European Conference on Computer Vision}}. Springer, \bibinfo{pages}{338--355}.
\newblock


\bibitem[Chen et~al\mbox{.}(2021)]%
        {chen2021mvsnerf}
\bibfield{author}{\bibinfo{person}{Anpei Chen}, \bibinfo{person}{Zexiang Xu}, \bibinfo{person}{Fuqiang Zhao}, \bibinfo{person}{Xiaoshuai Zhang}, \bibinfo{person}{Fanbo Xiang}, \bibinfo{person}{Jingyi Yu}, {and} \bibinfo{person}{Hao Su}.} \bibinfo{year}{2021}\natexlab{}.
\newblock \showarticletitle{Mvsnerf: Fast generalizable radiance field reconstruction from multi-view stereo}. In \bibinfo{booktitle}{\emph{Proceedings of the IEEE/CVF international conference on computer vision}}. \bibinfo{pages}{14124--14133}.
\newblock


\bibitem[Chen et~al\mbox{.}(2024)]%
        {chen2024mvdrag3d}
\bibfield{author}{\bibinfo{person}{Honghua Chen}, \bibinfo{person}{Yushi Lan}, \bibinfo{person}{Yongwei Chen}, \bibinfo{person}{Yifan Zhou}, {and} \bibinfo{person}{Xingang Pan}.} \bibinfo{year}{2024}\natexlab{}.
\newblock \showarticletitle{MvDrag3D: Drag-based Creative 3D Editing via Multi-view Generation-Reconstruction Priors}.
\newblock \bibinfo{journal}{\emph{arXiv preprint arXiv:2410.16272}} (\bibinfo{year}{2024}).
\newblock


\bibitem[Community(2018)]%
        {blender}
\bibfield{author}{\bibinfo{person}{Blender~Online Community}.} \bibinfo{year}{2018}\natexlab{}.
\newblock \bibinfo{booktitle}{\emph{Blender - a 3D modelling and rendering package}}.
\newblock Blender Foundation, Stichting Blender Foundation, Amsterdam.
\newblock
\urldef\tempurl%
\url{http://www.blender.org}
\showURL{%
\tempurl}


\bibitem[Deitke et~al\mbox{.}(2022)]%
        {objaverse}
\bibfield{author}{\bibinfo{person}{Matt Deitke}, \bibinfo{person}{Dustin Schwenk}, \bibinfo{person}{Jordi Salvador}, \bibinfo{person}{Luca Weihs}, \bibinfo{person}{Oscar Michel}, \bibinfo{person}{Eli VanderBilt}, \bibinfo{person}{Ludwig Schmidt}, \bibinfo{person}{Kiana Ehsani}, \bibinfo{person}{Aniruddha Kembhavi}, {and} \bibinfo{person}{Ali Farhadi}.} \bibinfo{year}{2022}\natexlab{}.
\newblock \showarticletitle{Objaverse: A Universe of Annotated 3D Objects}.
\newblock \bibinfo{journal}{\emph{arXiv preprint arXiv:2212.08051}} (\bibinfo{year}{2022}).
\newblock


\bibitem[Deng et~al\mbox{.}(2024)]%
        {deng2024detailgen3d}
\bibfield{author}{\bibinfo{person}{Ken Deng}, \bibinfo{person}{Yuanchen Guo}, \bibinfo{person}{Jingxiang Sun}, \bibinfo{person}{Zixin Zou}, \bibinfo{person}{Yangguang Li}, \bibinfo{person}{Xin Cai}, \bibinfo{person}{Yanpei Cao}, \bibinfo{person}{Yebin Liu}, {and} \bibinfo{person}{Ding Liang}.} \bibinfo{year}{2024}\natexlab{}.
\newblock \showarticletitle{DetailGen3D: Generative 3D Geometry Enhancement via Data-Dependent Flow}.
\newblock \bibinfo{journal}{\emph{arXiv preprint arXiv:2411.16820}} (\bibinfo{year}{2024}).
\newblock


\bibitem[Dong et~al\mbox{.}(2024)]%
        {dong2024interactive3d}
\bibfield{author}{\bibinfo{person}{Shaocong Dong}, \bibinfo{person}{Lihe Ding}, \bibinfo{person}{Zhanpeng Huang}, \bibinfo{person}{Zibin Wang}, \bibinfo{person}{Tianfan Xue}, {and} \bibinfo{person}{Dan Xu}.} \bibinfo{year}{2024}\natexlab{}.
\newblock \showarticletitle{Interactive3D: Create What You Want by Interactive 3D Generation}. In \bibinfo{booktitle}{\emph{Proceedings of the IEEE/CVF Conference on Computer Vision and Pattern Recognition}}. \bibinfo{pages}{4999--5008}.
\newblock


\bibitem[Downs et~al\mbox{.}(2022)]%
        {downs2022google}
\bibfield{author}{\bibinfo{person}{Laura Downs}, \bibinfo{person}{Anthony Francis}, \bibinfo{person}{Nate Koenig}, \bibinfo{person}{Brandon Kinman}, \bibinfo{person}{Ryan Hickman}, \bibinfo{person}{Krista Reymann}, \bibinfo{person}{Thomas~B McHugh}, {and} \bibinfo{person}{Vincent Vanhoucke}.} \bibinfo{year}{2022}\natexlab{}.
\newblock \showarticletitle{Google scanned objects: A high-quality dataset of 3d scanned household items}. In \bibinfo{booktitle}{\emph{2022 International Conference on Robotics and Automation (ICRA)}}. IEEE, \bibinfo{pages}{2553--2560}.
\newblock


\bibitem[Fischer et~al\mbox{.}(2023)]%
        {fischer2023boosting}
\bibfield{author}{\bibinfo{person}{Johannes~S Fischer}, \bibinfo{person}{Ming Gui}, \bibinfo{person}{Pingchuan Ma}, \bibinfo{person}{Nick Stracke}, \bibinfo{person}{Stefan~A Baumann}, {and} \bibinfo{person}{Bj{\"o}rn Ommer}.} \bibinfo{year}{2023}\natexlab{}.
\newblock \showarticletitle{Boosting Latent Diffusion with Flow Matching}.
\newblock \bibinfo{journal}{\emph{arXiv preprint arXiv:2312.07360}} (\bibinfo{year}{2023}).
\newblock


\bibitem[Fridovich-Keil et~al\mbox{.}(2022)]%
        {fridovich2022plenoxels}
\bibfield{author}{\bibinfo{person}{Sara Fridovich-Keil}, \bibinfo{person}{Alex Yu}, \bibinfo{person}{Matthew Tancik}, \bibinfo{person}{Qinhong Chen}, \bibinfo{person}{Benjamin Recht}, {and} \bibinfo{person}{Angjoo Kanazawa}.} \bibinfo{year}{2022}\natexlab{}.
\newblock \showarticletitle{Plenoxels: Radiance fields without neural networks}. In \bibinfo{booktitle}{\emph{Proceedings of the IEEE/CVF conference on computer vision and pattern recognition}}. \bibinfo{pages}{5501--5510}.
\newblock


\bibitem[Gupta et~al\mbox{.}(2023)]%
        {gupta20233dgen}
\bibfield{author}{\bibinfo{person}{Anchit Gupta}, \bibinfo{person}{Wenhan Xiong}, \bibinfo{person}{Yixin Nie}, \bibinfo{person}{Ian Jones}, {and} \bibinfo{person}{Barlas O{\u{g}}uz}.} \bibinfo{year}{2023}\natexlab{}.
\newblock \showarticletitle{3dgen: Triplane latent diffusion for textured mesh generation}.
\newblock \bibinfo{journal}{\emph{arXiv preprint arXiv:2303.05371}} (\bibinfo{year}{2023}).
\newblock


\bibitem[He et~al\mbox{.}(2025)]%
        {he2025gvgen}
\bibfield{author}{\bibinfo{person}{Xianglong He}, \bibinfo{person}{Junyi Chen}, \bibinfo{person}{Sida Peng}, \bibinfo{person}{Di Huang}, \bibinfo{person}{Yangguang Li}, \bibinfo{person}{Xiaoshui Huang}, \bibinfo{person}{Chun Yuan}, \bibinfo{person}{Wanli Ouyang}, {and} \bibinfo{person}{Tong He}.} \bibinfo{year}{2025}\natexlab{}.
\newblock \showarticletitle{Gvgen: Text-to-3d generation with volumetric representation}. In \bibinfo{booktitle}{\emph{European Conference on Computer Vision}}. Springer, \bibinfo{pages}{463--479}.
\newblock


\bibitem[Hedman et~al\mbox{.}(2021)]%
        {hedman2021snerg}
\bibfield{author}{\bibinfo{person}{Peter Hedman}, \bibinfo{person}{Pratul~P. Srinivasan}, \bibinfo{person}{Ben Mildenhall}, \bibinfo{person}{Jonathan~T. Barron}, {and} \bibinfo{person}{Paul Debevec}.} \bibinfo{year}{2021}\natexlab{}.
\newblock \showarticletitle{Baking Neural Radiance Fields for Real-Time View Synthesis}.
\newblock \bibinfo{journal}{\emph{ICCV}} (\bibinfo{year}{2021}).
\newblock


\bibitem[Hong et~al\mbox{.}(2023)]%
        {hong2023lrm}
\bibfield{author}{\bibinfo{person}{Yicong Hong}, \bibinfo{person}{Kai Zhang}, \bibinfo{person}{Jiuxiang Gu}, \bibinfo{person}{Sai Bi}, \bibinfo{person}{Yang Zhou}, \bibinfo{person}{Difan Liu}, \bibinfo{person}{Feng Liu}, \bibinfo{person}{Kalyan Sunkavalli}, \bibinfo{person}{Trung Bui}, {and} \bibinfo{person}{Hao Tan}.} \bibinfo{year}{2023}\natexlab{}.
\newblock \showarticletitle{Lrm: Large reconstruction model for single image to 3d}.
\newblock \bibinfo{journal}{\emph{arXiv preprint arXiv:2311.04400}} (\bibinfo{year}{2023}).
\newblock


\bibitem[Huang et~al\mbox{.}(2024)]%
        {huang20242d}
\bibfield{author}{\bibinfo{person}{Binbin Huang}, \bibinfo{person}{Zehao Yu}, \bibinfo{person}{Anpei Chen}, \bibinfo{person}{Andreas Geiger}, {and} \bibinfo{person}{Shenghua Gao}.} \bibinfo{year}{2024}\natexlab{}.
\newblock \showarticletitle{2d gaussian splatting for geometrically accurate radiance fields}. In \bibinfo{booktitle}{\emph{ACM SIGGRAPH 2024 Conference Papers}}. \bibinfo{pages}{1--11}.
\newblock


\bibitem[Johari et~al\mbox{.}(2022)]%
        {johari2022geonerf}
\bibfield{author}{\bibinfo{person}{Mohammad~Mahdi Johari}, \bibinfo{person}{Yann Lepoittevin}, {and} \bibinfo{person}{Fran{\c{c}}ois Fleuret}.} \bibinfo{year}{2022}\natexlab{}.
\newblock \showarticletitle{Geonerf: Generalizing nerf with geometry priors}. In \bibinfo{booktitle}{\emph{Proceedings of the IEEE/CVF Conference on Computer Vision and Pattern Recognition}}. \bibinfo{pages}{18365--18375}.
\newblock


\bibitem[Kerbl et~al\mbox{.}(2023)]%
        {kerbl3Dgaussians}
\bibfield{author}{\bibinfo{person}{Bernhard Kerbl}, \bibinfo{person}{Georgios Kopanas}, \bibinfo{person}{Thomas Leimk{\"u}hler}, {and} \bibinfo{person}{George Drettakis}.} \bibinfo{year}{2023}\natexlab{}.
\newblock \showarticletitle{3D Gaussian Splatting for Real-Time Radiance Field Rendering}.
\newblock \bibinfo{journal}{\emph{ACM Transactions on Graphics}} \bibinfo{volume}{42}, \bibinfo{number}{4} (\bibinfo{date}{July} \bibinfo{year}{2023}).
\newblock
\urldef\tempurl%
\url{https://repo-sam.inria.fr/fungraph/3d-gaussian-splatting/}
\showURL{%
\tempurl}


\bibitem[Li et~al\mbox{.}(2023b)]%
        {li2023instant3d}
\bibfield{author}{\bibinfo{person}{Jiahao Li}, \bibinfo{person}{Hao Tan}, \bibinfo{person}{Kai Zhang}, \bibinfo{person}{Zexiang Xu}, \bibinfo{person}{Fujun Luan}, \bibinfo{person}{Yinghao Xu}, \bibinfo{person}{Yicong Hong}, \bibinfo{person}{Kalyan Sunkavalli}, \bibinfo{person}{Greg Shakhnarovich}, {and} \bibinfo{person}{Sai Bi}.} \bibinfo{year}{2023}\natexlab{b}.
\newblock \showarticletitle{Instant3d: Fast text-to-3d with sparse-view generation and large reconstruction model}.
\newblock \bibinfo{journal}{\emph{arXiv preprint arXiv:2311.06214}} (\bibinfo{year}{2023}).
\newblock


\bibitem[Li et~al\mbox{.}(2023a)]%
        {li2023sweetdreamer}
\bibfield{author}{\bibinfo{person}{Weiyu Li}, \bibinfo{person}{Rui Chen}, \bibinfo{person}{Xuelin Chen}, {and} \bibinfo{person}{Ping Tan}.} \bibinfo{year}{2023}\natexlab{a}.
\newblock \showarticletitle{Sweetdreamer: Aligning geometric priors in 2d diffusion for consistent text-to-3d}.
\newblock \bibinfo{journal}{\emph{arXiv preprint arXiv:2310.02596}} (\bibinfo{year}{2023}).
\newblock


\bibitem[Liang et~al\mbox{.}(2024)]%
        {liang2024luciddreamer}
\bibfield{author}{\bibinfo{person}{Yixun Liang}, \bibinfo{person}{Xin Yang}, \bibinfo{person}{Jiantao Lin}, \bibinfo{person}{Haodong Li}, \bibinfo{person}{Xiaogang Xu}, {and} \bibinfo{person}{Yingcong Chen}.} \bibinfo{year}{2024}\natexlab{}.
\newblock \showarticletitle{Luciddreamer: Towards high-fidelity text-to-3d generation via interval score matching}. In \bibinfo{booktitle}{\emph{Proceedings of the IEEE/CVF Conference on Computer Vision and Pattern Recognition}}. \bibinfo{pages}{6517--6526}.
\newblock


\bibitem[Lipman et~al\mbox{.}(2022)]%
        {lipman2022flow}
\bibfield{author}{\bibinfo{person}{Yaron Lipman}, \bibinfo{person}{Ricky~TQ Chen}, \bibinfo{person}{Heli Ben-Hamu}, \bibinfo{person}{Maximilian Nickel}, {and} \bibinfo{person}{Matt Le}.} \bibinfo{year}{2022}\natexlab{}.
\newblock \showarticletitle{Flow matching for generative modeling}.
\newblock \bibinfo{journal}{\emph{arXiv preprint arXiv:2210.02747}} (\bibinfo{year}{2022}).
\newblock


\bibitem[Liu et~al\mbox{.}(2024a)]%
        {liu2024sketchdream}
\bibfield{author}{\bibinfo{person}{Feng-Lin Liu}, \bibinfo{person}{Hongbo Fu}, \bibinfo{person}{Yu-Kun Lai}, {and} \bibinfo{person}{Lin Gao}.} \bibinfo{year}{2024}\natexlab{a}.
\newblock \showarticletitle{Sketchdream: Sketch-based text-to-3d generation and editing}.
\newblock \bibinfo{journal}{\emph{ACM Transactions on Graphics (TOG)}} \bibinfo{volume}{43}, \bibinfo{number}{4} (\bibinfo{year}{2024}), \bibinfo{pages}{1--13}.
\newblock


\bibitem[Liu et~al\mbox{.}(2024b)]%
        {liu2024one++}
\bibfield{author}{\bibinfo{person}{Minghua Liu}, \bibinfo{person}{Ruoxi Shi}, \bibinfo{person}{Linghao Chen}, \bibinfo{person}{Zhuoyang Zhang}, \bibinfo{person}{Chao Xu}, \bibinfo{person}{Xinyue Wei}, \bibinfo{person}{Hansheng Chen}, \bibinfo{person}{Chong Zeng}, \bibinfo{person}{Jiayuan Gu}, {and} \bibinfo{person}{Hao Su}.} \bibinfo{year}{2024}\natexlab{b}.
\newblock \showarticletitle{One-2-3-45++: Fast single image to 3d objects with consistent multi-view generation and 3d diffusion}. In \bibinfo{booktitle}{\emph{Proceedings of the IEEE/CVF Conference on Computer Vision and Pattern Recognition}}. \bibinfo{pages}{10072--10083}.
\newblock


\bibitem[Liu et~al\mbox{.}(2024c)]%
        {liu2024one}
\bibfield{author}{\bibinfo{person}{Minghua Liu}, \bibinfo{person}{Chao Xu}, \bibinfo{person}{Haian Jin}, \bibinfo{person}{Linghao Chen}, \bibinfo{person}{Mukund Varma~T}, \bibinfo{person}{Zexiang Xu}, {and} \bibinfo{person}{Hao Su}.} \bibinfo{year}{2024}\natexlab{c}.
\newblock \showarticletitle{One-2-3-45: Any single image to 3d mesh in 45 seconds without per-shape optimization}.
\newblock \bibinfo{journal}{\emph{Advances in Neural Information Processing Systems}}  \bibinfo{volume}{36} (\bibinfo{year}{2024}).
\newblock


\bibitem[Liu et~al\mbox{.}(2023)]%
        {liu2023zero}
\bibfield{author}{\bibinfo{person}{Ruoshi Liu}, \bibinfo{person}{Rundi Wu}, \bibinfo{person}{Basile Van~Hoorick}, \bibinfo{person}{Pavel Tokmakov}, \bibinfo{person}{Sergey Zakharov}, {and} \bibinfo{person}{Carl Vondrick}.} \bibinfo{year}{2023}\natexlab{}.
\newblock \showarticletitle{Zero-1-to-3: Zero-shot one image to 3d object}. In \bibinfo{booktitle}{\emph{Proceedings of the IEEE/CVF international conference on computer vision}}. \bibinfo{pages}{9298--9309}.
\newblock


\bibitem[Liu et~al\mbox{.}(2022)]%
        {liu2022flow}
\bibfield{author}{\bibinfo{person}{Xingchao Liu}, \bibinfo{person}{Chengyue Gong}, {and} \bibinfo{person}{Qiang Liu}.} \bibinfo{year}{2022}\natexlab{}.
\newblock \showarticletitle{Flow straight and fast: Learning to generate and transfer data with rectified flow}.
\newblock \bibinfo{journal}{\emph{arXiv preprint arXiv:2209.03003}} (\bibinfo{year}{2022}).
\newblock


\bibitem[Long et~al\mbox{.}(2024)]%
        {long2024wonder3d}
\bibfield{author}{\bibinfo{person}{Xiaoxiao Long}, \bibinfo{person}{Yuan-Chen Guo}, \bibinfo{person}{Cheng Lin}, \bibinfo{person}{Yuan Liu}, \bibinfo{person}{Zhiyang Dou}, \bibinfo{person}{Lingjie Liu}, \bibinfo{person}{Yuexin Ma}, \bibinfo{person}{Song-Hai Zhang}, \bibinfo{person}{Marc Habermann}, \bibinfo{person}{Christian Theobalt}, {et~al\mbox{.}}} \bibinfo{year}{2024}\natexlab{}.
\newblock \showarticletitle{Wonder3d: Single image to 3d using cross-domain diffusion}. In \bibinfo{booktitle}{\emph{Proceedings of the IEEE/CVF Conference on Computer Vision and Pattern Recognition}}. \bibinfo{pages}{9970--9980}.
\newblock


\bibitem[Lu et~al\mbox{.}(2024)]%
        {lu2024scaffold}
\bibfield{author}{\bibinfo{person}{Tao Lu}, \bibinfo{person}{Mulin Yu}, \bibinfo{person}{Linning Xu}, \bibinfo{person}{Yuanbo Xiangli}, \bibinfo{person}{Limin Wang}, \bibinfo{person}{Dahua Lin}, {and} \bibinfo{person}{Bo Dai}.} \bibinfo{year}{2024}\natexlab{}.
\newblock \showarticletitle{Scaffold-gs: Structured 3d gaussians for view-adaptive rendering}. In \bibinfo{booktitle}{\emph{Proceedings of the IEEE/CVF Conference on Computer Vision and Pattern Recognition}}. \bibinfo{pages}{20654--20664}.
\newblock


\bibitem[Mildenhall et~al\mbox{.}(2021)]%
        {mildenhall2021nerf}
\bibfield{author}{\bibinfo{person}{Ben Mildenhall}, \bibinfo{person}{Pratul~P Srinivasan}, \bibinfo{person}{Matthew Tancik}, \bibinfo{person}{Jonathan~T Barron}, \bibinfo{person}{Ravi Ramamoorthi}, {and} \bibinfo{person}{Ren Ng}.} \bibinfo{year}{2021}\natexlab{}.
\newblock \showarticletitle{Nerf: Representing scenes as neural radiance fields for view synthesis}.
\newblock \bibinfo{journal}{\emph{Commun. ACM}} \bibinfo{volume}{65}, \bibinfo{number}{1} (\bibinfo{year}{2021}), \bibinfo{pages}{99--106}.
\newblock


\bibitem[M{\"u}ller et~al\mbox{.}(2023)]%
        {muller2023diffrf}
\bibfield{author}{\bibinfo{person}{Norman M{\"u}ller}, \bibinfo{person}{Yawar Siddiqui}, \bibinfo{person}{Lorenzo Porzi}, \bibinfo{person}{Samuel~Rota Bulo}, \bibinfo{person}{Peter Kontschieder}, {and} \bibinfo{person}{Matthias Nie{\ss}ner}.} \bibinfo{year}{2023}\natexlab{}.
\newblock \showarticletitle{Diffrf: Rendering-guided 3d radiance field diffusion}. In \bibinfo{booktitle}{\emph{Proceedings of the IEEE/CVF Conference on Computer Vision and Pattern Recognition}}. \bibinfo{pages}{4328--4338}.
\newblock


\bibitem[M{\"u}ller et~al\mbox{.}(2022)]%
        {muller2022instant}
\bibfield{author}{\bibinfo{person}{Thomas M{\"u}ller}, \bibinfo{person}{Alex Evans}, \bibinfo{person}{Christoph Schied}, {and} \bibinfo{person}{Alexander Keller}.} \bibinfo{year}{2022}\natexlab{}.
\newblock \showarticletitle{Instant neural graphics primitives with a multiresolution hash encoding}.
\newblock \bibinfo{journal}{\emph{ACM transactions on graphics (TOG)}} \bibinfo{volume}{41}, \bibinfo{number}{4} (\bibinfo{year}{2022}), \bibinfo{pages}{1--15}.
\newblock


\bibitem[Nichol et~al\mbox{.}(2022)]%
        {nichol2022point}
\bibfield{author}{\bibinfo{person}{Alex Nichol}, \bibinfo{person}{Heewoo Jun}, \bibinfo{person}{Prafulla Dhariwal}, \bibinfo{person}{Pamela Mishkin}, {and} \bibinfo{person}{Mark Chen}.} \bibinfo{year}{2022}\natexlab{}.
\newblock \showarticletitle{Point-e: A system for generating 3d point clouds from complex prompts}.
\newblock \bibinfo{journal}{\emph{arXiv preprint arXiv:2212.08751}} (\bibinfo{year}{2022}).
\newblock


\bibitem[Oquab et~al\mbox{.}(2023)]%
        {oquab2023dinov2}
\bibfield{author}{\bibinfo{person}{Maxime Oquab}, \bibinfo{person}{Timoth{\'e}e Darcet}, \bibinfo{person}{Th{\'e}o Moutakanni}, \bibinfo{person}{Huy Vo}, \bibinfo{person}{Marc Szafraniec}, \bibinfo{person}{Vasil Khalidov}, \bibinfo{person}{Pierre Fernandez}, \bibinfo{person}{Daniel Haziza}, \bibinfo{person}{Francisco Massa}, \bibinfo{person}{Alaaeldin El-Nouby}, {et~al\mbox{.}}} \bibinfo{year}{2023}\natexlab{}.
\newblock \showarticletitle{Dinov2: Learning robust visual features without supervision}.
\newblock \bibinfo{journal}{\emph{arXiv preprint arXiv:2304.07193}} (\bibinfo{year}{2023}).
\newblock


\bibitem[Peebles and Xie(2023)]%
        {peebles2023scalable}
\bibfield{author}{\bibinfo{person}{William Peebles} {and} \bibinfo{person}{Saining Xie}.} \bibinfo{year}{2023}\natexlab{}.
\newblock \showarticletitle{Scalable diffusion models with transformers}. In \bibinfo{booktitle}{\emph{Proceedings of the IEEE/CVF International Conference on Computer Vision}}. \bibinfo{pages}{4195--4205}.
\newblock


\bibitem[Poole et~al\mbox{.}(2022)]%
        {poole2022dreamfusion}
\bibfield{author}{\bibinfo{person}{Ben Poole}, \bibinfo{person}{Ajay Jain}, \bibinfo{person}{Jonathan~T Barron}, {and} \bibinfo{person}{Ben Mildenhall}.} \bibinfo{year}{2022}\natexlab{}.
\newblock \showarticletitle{Dreamfusion: Text-to-3d using 2d diffusion}.
\newblock \bibinfo{journal}{\emph{arXiv preprint arXiv:2209.14988}} (\bibinfo{year}{2022}).
\newblock


\bibitem[Qiu et~al\mbox{.}(2023)]%
        {qiu2023richdreamer}
\bibfield{author}{\bibinfo{person}{Lingteng Qiu}, \bibinfo{person}{Guanying Chen}, \bibinfo{person}{Xiaodong Gu}, \bibinfo{person}{Qi zuo}, \bibinfo{person}{Mutian Xu}, \bibinfo{person}{Yushuang Wu}, \bibinfo{person}{Weihao Yuan}, \bibinfo{person}{Zilong Dong}, \bibinfo{person}{Liefeng Bo}, {and} \bibinfo{person}{Xiaoguang Han}.} \bibinfo{year}{2023}\natexlab{}.
\newblock \showarticletitle{RichDreamer: A Generalizable Normal-Depth Diffusion Model for Detail Richness in Text-to-3D}.
\newblock \bibinfo{journal}{\emph{arXiv preprint arXiv:2311.16918}} (\bibinfo{year}{2023}).
\newblock


\bibitem[Rombach et~al\mbox{.}(2022)]%
        {rombach2022high}
\bibfield{author}{\bibinfo{person}{Robin Rombach}, \bibinfo{person}{Andreas Blattmann}, \bibinfo{person}{Dominik Lorenz}, \bibinfo{person}{Patrick Esser}, {and} \bibinfo{person}{Bj{\"o}rn Ommer}.} \bibinfo{year}{2022}\natexlab{}.
\newblock \showarticletitle{High-resolution image synthesis with latent diffusion models}. In \bibinfo{booktitle}{\emph{Proceedings of the IEEE/CVF conference on computer vision and pattern recognition}}. \bibinfo{pages}{10684--10695}.
\newblock


\bibitem[Saharia et~al\mbox{.}(2022)]%
        {saharia2022photorealistic}
\bibfield{author}{\bibinfo{person}{Chitwan Saharia}, \bibinfo{person}{William Chan}, \bibinfo{person}{Saurabh Saxena}, \bibinfo{person}{Lala Li}, \bibinfo{person}{Jay Whang}, \bibinfo{person}{Emily~L Denton}, \bibinfo{person}{Kamyar Ghasemipour}, \bibinfo{person}{Raphael Gontijo~Lopes}, \bibinfo{person}{Burcu Karagol~Ayan}, \bibinfo{person}{Tim Salimans}, {et~al\mbox{.}}} \bibinfo{year}{2022}\natexlab{}.
\newblock \showarticletitle{Photorealistic text-to-image diffusion models with deep language understanding}.
\newblock \bibinfo{journal}{\emph{Advances in neural information processing systems}}  \bibinfo{volume}{35} (\bibinfo{year}{2022}), \bibinfo{pages}{36479--36494}.
\newblock


\bibitem[Shi et~al\mbox{.}(2023a)]%
        {shi2023zero123++}
\bibfield{author}{\bibinfo{person}{Ruoxi Shi}, \bibinfo{person}{Hansheng Chen}, \bibinfo{person}{Zhuoyang Zhang}, \bibinfo{person}{Minghua Liu}, \bibinfo{person}{Chao Xu}, \bibinfo{person}{Xinyue Wei}, \bibinfo{person}{Linghao Chen}, \bibinfo{person}{Chong Zeng}, {and} \bibinfo{person}{Hao Su}.} \bibinfo{year}{2023}\natexlab{a}.
\newblock \showarticletitle{Zero123++: a single image to consistent multi-view diffusion base model}.
\newblock \bibinfo{journal}{\emph{arXiv preprint arXiv:2310.15110}} (\bibinfo{year}{2023}).
\newblock


\bibitem[Shi et~al\mbox{.}(2023b)]%
        {shi2023mvdream}
\bibfield{author}{\bibinfo{person}{Yichun Shi}, \bibinfo{person}{Peng Wang}, \bibinfo{person}{Jianglong Ye}, \bibinfo{person}{Mai Long}, \bibinfo{person}{Kejie Li}, {and} \bibinfo{person}{Xiao Yang}.} \bibinfo{year}{2023}\natexlab{b}.
\newblock \showarticletitle{Mvdream: Multi-view diffusion for 3d generation}.
\newblock \bibinfo{journal}{\emph{arXiv preprint arXiv:2308.16512}} (\bibinfo{year}{2023}).
\newblock


\bibitem[Sun et~al\mbox{.}(2022)]%
        {SunSC22}
\bibfield{author}{\bibinfo{person}{Cheng Sun}, \bibinfo{person}{Min Sun}, {and} \bibinfo{person}{Hwann{-}Tzong Chen}.} \bibinfo{year}{2022}\natexlab{}.
\newblock \showarticletitle{Direct Voxel Grid Optimization: Super-fast Convergence for Radiance Fields Reconstruction}. In \bibinfo{booktitle}{\emph{CVPR}}.
\newblock


\bibitem[Tang et~al\mbox{.}(2025a)]%
        {tang2025lgm}
\bibfield{author}{\bibinfo{person}{Jiaxiang Tang}, \bibinfo{person}{Zhaoxi Chen}, \bibinfo{person}{Xiaokang Chen}, \bibinfo{person}{Tengfei Wang}, \bibinfo{person}{Gang Zeng}, {and} \bibinfo{person}{Ziwei Liu}.} \bibinfo{year}{2025}\natexlab{a}.
\newblock \showarticletitle{Lgm: Large multi-view gaussian model for high-resolution 3d content creation}. In \bibinfo{booktitle}{\emph{European Conference on Computer Vision}}. Springer, \bibinfo{pages}{1--18}.
\newblock


\bibitem[Tang et~al\mbox{.}(2023)]%
        {tang2023dreamgaussian}
\bibfield{author}{\bibinfo{person}{Jiaxiang Tang}, \bibinfo{person}{Jiawei Ren}, \bibinfo{person}{Hang Zhou}, \bibinfo{person}{Ziwei Liu}, {and} \bibinfo{person}{Gang Zeng}.} \bibinfo{year}{2023}\natexlab{}.
\newblock \showarticletitle{Dreamgaussian: Generative gaussian splatting for efficient 3d content creation}.
\newblock \bibinfo{journal}{\emph{arXiv preprint arXiv:2309.16653}} (\bibinfo{year}{2023}).
\newblock


\bibitem[Tang et~al\mbox{.}(2025b)]%
        {tang2025mvdiffusion++}
\bibfield{author}{\bibinfo{person}{Shitao Tang}, \bibinfo{person}{Jiacheng Chen}, \bibinfo{person}{Dilin Wang}, \bibinfo{person}{Chengzhou Tang}, \bibinfo{person}{Fuyang Zhang}, \bibinfo{person}{Yuchen Fan}, \bibinfo{person}{Vikas Chandra}, \bibinfo{person}{Yasutaka Furukawa}, {and} \bibinfo{person}{Rakesh Ranjan}.} \bibinfo{year}{2025}\natexlab{b}.
\newblock \showarticletitle{Mvdiffusion++: A dense high-resolution multi-view diffusion model for single or sparse-view 3d object reconstruction}. In \bibinfo{booktitle}{\emph{European Conference on Computer Vision}}. Springer, \bibinfo{pages}{175--191}.
\newblock


\bibitem[Tochilkin et~al\mbox{.}(2024)]%
        {tochilkin2024triposr}
\bibfield{author}{\bibinfo{person}{Dmitry Tochilkin}, \bibinfo{person}{David Pankratz}, \bibinfo{person}{Zexiang Liu}, \bibinfo{person}{Zixuan Huang}, \bibinfo{person}{Adam Letts}, \bibinfo{person}{Yangguang Li}, \bibinfo{person}{Ding Liang}, \bibinfo{person}{Christian Laforte}, \bibinfo{person}{Varun Jampani}, {and} \bibinfo{person}{Yan-Pei Cao}.} \bibinfo{year}{2024}\natexlab{}.
\newblock \showarticletitle{Triposr: Fast 3d object reconstruction from a single image}.
\newblock \bibinfo{journal}{\emph{arXiv preprint arXiv:2403.02151}} (\bibinfo{year}{2024}).
\newblock


\bibitem[Wang and Shi(2023)]%
        {wang2023imagedream}
\bibfield{author}{\bibinfo{person}{Peng Wang} {and} \bibinfo{person}{Yichun Shi}.} \bibinfo{year}{2023}\natexlab{}.
\newblock \showarticletitle{Imagedream: Image-prompt multi-view diffusion for 3d generation}.
\newblock \bibinfo{journal}{\emph{arXiv preprint arXiv:2312.02201}} (\bibinfo{year}{2023}).
\newblock


\bibitem[Wang et~al\mbox{.}(2021)]%
        {wang2021ibrnet}
\bibfield{author}{\bibinfo{person}{Qianqian Wang}, \bibinfo{person}{Zhicheng Wang}, \bibinfo{person}{Kyle Genova}, \bibinfo{person}{Pratul~P Srinivasan}, \bibinfo{person}{Howard Zhou}, \bibinfo{person}{Jonathan~T Barron}, \bibinfo{person}{Ricardo Martin-Brualla}, \bibinfo{person}{Noah Snavely}, {and} \bibinfo{person}{Thomas Funkhouser}.} \bibinfo{year}{2021}\natexlab{}.
\newblock \showarticletitle{Ibrnet: Learning multi-view image-based rendering}. In \bibinfo{booktitle}{\emph{Proceedings of the IEEE/CVF conference on computer vision and pattern recognition}}. \bibinfo{pages}{4690--4699}.
\newblock


\bibitem[Wang et~al\mbox{.}(2024)]%
        {wang2024prolificdreamer}
\bibfield{author}{\bibinfo{person}{Zhengyi Wang}, \bibinfo{person}{Cheng Lu}, \bibinfo{person}{Yikai Wang}, \bibinfo{person}{Fan Bao}, \bibinfo{person}{Chongxuan Li}, \bibinfo{person}{Hang Su}, {and} \bibinfo{person}{Jun Zhu}.} \bibinfo{year}{2024}\natexlab{}.
\newblock \showarticletitle{Prolificdreamer: High-fidelity and diverse text-to-3d generation with variational score distillation}.
\newblock \bibinfo{journal}{\emph{Advances in Neural Information Processing Systems}}  \bibinfo{volume}{36} (\bibinfo{year}{2024}).
\newblock


\bibitem[Wang et~al\mbox{.}(2025)]%
        {wang2025crm}
\bibfield{author}{\bibinfo{person}{Zhengyi Wang}, \bibinfo{person}{Yikai Wang}, \bibinfo{person}{Yifei Chen}, \bibinfo{person}{Chendong Xiang}, \bibinfo{person}{Shuo Chen}, \bibinfo{person}{Dajiang Yu}, \bibinfo{person}{Chongxuan Li}, \bibinfo{person}{Hang Su}, {and} \bibinfo{person}{Jun Zhu}.} \bibinfo{year}{2025}\natexlab{}.
\newblock \showarticletitle{Crm: Single image to 3d textured mesh with convolutional reconstruction model}. In \bibinfo{booktitle}{\emph{European Conference on Computer Vision}}. Springer, \bibinfo{pages}{57--74}.
\newblock


\bibitem[Xiang et~al\mbox{.}(2024)]%
        {xiang2024structured}
\bibfield{author}{\bibinfo{person}{Jianfeng Xiang}, \bibinfo{person}{Zelong Lv}, \bibinfo{person}{Sicheng Xu}, \bibinfo{person}{Yu Deng}, \bibinfo{person}{Ruicheng Wang}, \bibinfo{person}{Bowen Zhang}, \bibinfo{person}{Dong Chen}, \bibinfo{person}{Xin Tong}, {and} \bibinfo{person}{Jiaolong Yang}.} \bibinfo{year}{2024}\natexlab{}.
\newblock \showarticletitle{Structured 3D Latents for Scalable and Versatile 3D Generation}.
\newblock \bibinfo{journal}{\emph{arXiv preprint arXiv:2412.01506}} (\bibinfo{year}{2024}).
\newblock


\bibitem[Xu et~al\mbox{.}(2024a)]%
        {xu2024instantmesh}
\bibfield{author}{\bibinfo{person}{Jiale Xu}, \bibinfo{person}{Weihao Cheng}, \bibinfo{person}{Yiming Gao}, \bibinfo{person}{Xintao Wang}, \bibinfo{person}{Shenghua Gao}, {and} \bibinfo{person}{Ying Shan}.} \bibinfo{year}{2024}\natexlab{a}.
\newblock \showarticletitle{Instantmesh: Efficient 3d mesh generation from a single image with sparse-view large reconstruction models}.
\newblock \bibinfo{journal}{\emph{arXiv preprint arXiv:2404.07191}} (\bibinfo{year}{2024}).
\newblock


\bibitem[Xu et~al\mbox{.}(2024b)]%
        {xu2024grm}
\bibfield{author}{\bibinfo{person}{Yinghao Xu}, \bibinfo{person}{Zifan Shi}, \bibinfo{person}{Wang Yifan}, \bibinfo{person}{Hansheng Chen}, \bibinfo{person}{Ceyuan Yang}, \bibinfo{person}{Sida Peng}, \bibinfo{person}{Yujun Shen}, {and} \bibinfo{person}{Gordon Wetzstein}.} \bibinfo{year}{2024}\natexlab{b}.
\newblock \showarticletitle{Grm: Large gaussian reconstruction model for efficient 3d reconstruction and generation}.
\newblock \bibinfo{journal}{\emph{arXiv preprint arXiv:2403.14621}} (\bibinfo{year}{2024}).
\newblock


\bibitem[Xu et~al\mbox{.}(2023)]%
        {xu2023dmv3d}
\bibfield{author}{\bibinfo{person}{Yinghao Xu}, \bibinfo{person}{Hao Tan}, \bibinfo{person}{Fujun Luan}, \bibinfo{person}{Sai Bi}, \bibinfo{person}{Peng Wang}, \bibinfo{person}{Jiahao Li}, \bibinfo{person}{Zifan Shi}, \bibinfo{person}{Kalyan Sunkavalli}, \bibinfo{person}{Gordon Wetzstein}, \bibinfo{person}{Zexiang Xu}, {et~al\mbox{.}}} \bibinfo{year}{2023}\natexlab{}.
\newblock \showarticletitle{Dmv3d: Denoising multi-view diffusion using 3d large reconstruction model}.
\newblock \bibinfo{journal}{\emph{arXiv preprint arXiv:2311.09217}} (\bibinfo{year}{2023}).
\newblock


\bibitem[Yu et~al\mbox{.}(2021)]%
        {yu2021pixelnerf}
\bibfield{author}{\bibinfo{person}{Alex Yu}, \bibinfo{person}{Vickie Ye}, \bibinfo{person}{Matthew Tancik}, {and} \bibinfo{person}{Angjoo Kanazawa}.} \bibinfo{year}{2021}\natexlab{}.
\newblock \showarticletitle{pixelnerf: Neural radiance fields from one or few images}. In \bibinfo{booktitle}{\emph{Proceedings of the IEEE/CVF conference on computer vision and pattern recognition}}. \bibinfo{pages}{4578--4587}.
\newblock


\bibitem[Yu et~al\mbox{.}(2024a)]%
        {yu2024mip}
\bibfield{author}{\bibinfo{person}{Zehao Yu}, \bibinfo{person}{Anpei Chen}, \bibinfo{person}{Binbin Huang}, \bibinfo{person}{Torsten Sattler}, {and} \bibinfo{person}{Andreas Geiger}.} \bibinfo{year}{2024}\natexlab{a}.
\newblock \showarticletitle{Mip-splatting: Alias-free 3d gaussian splatting}. In \bibinfo{booktitle}{\emph{Proceedings of the IEEE/CVF Conference on Computer Vision and Pattern Recognition}}. \bibinfo{pages}{19447--19456}.
\newblock


\bibitem[Yu et~al\mbox{.}(2024b)]%
        {Yu2024GOF}
\bibfield{author}{\bibinfo{person}{Zehao Yu}, \bibinfo{person}{Torsten Sattler}, {and} \bibinfo{person}{Andreas Geiger}.} \bibinfo{year}{2024}\natexlab{b}.
\newblock \showarticletitle{Gaussian Opacity Fields: Efficient Adaptive Surface Reconstruction in Unbounded Scenes}.
\newblock \bibinfo{journal}{\emph{ACM Transactions on Graphics}} (\bibinfo{year}{2024}).
\newblock


\bibitem[Yuksel(2015)]%
        {poisson}
\bibfield{author}{\bibinfo{person}{Cem Yuksel}.} \bibinfo{year}{2015}\natexlab{}.
\newblock \showarticletitle{Sample Elimination for Generating Poisson Disk Sample Sets}.
\newblock \bibinfo{journal}{\emph{Comput. Graph. Forum}} \bibinfo{volume}{34}, \bibinfo{number}{2} (\bibinfo{date}{May} \bibinfo{year}{2015}), \bibinfo{pages}{25–32}.
\newblock
\showISSN{0167-7055}
\urldef\tempurl%
\url{https://doi.org/10.1111/cgf.12538}
\showDOI{\tempurl}


\bibitem[Zhang et~al\mbox{.}(2024a)]%
        {zhang2024gaussiancube}
\bibfield{author}{\bibinfo{person}{Bowen Zhang}, \bibinfo{person}{Yiji Cheng}, \bibinfo{person}{Jiaolong Yang}, \bibinfo{person}{Chunyu Wang}, \bibinfo{person}{Feng Zhao}, \bibinfo{person}{Yansong Tang}, \bibinfo{person}{Dong Chen}, {and} \bibinfo{person}{Baining Guo}.} \bibinfo{year}{2024}\natexlab{a}.
\newblock \showarticletitle{GaussianCube: Structuring Gaussian Splatting using Optimal Transport for 3D Generative Modeling}.
\newblock \bibinfo{journal}{\emph{arXiv preprint arXiv:2403.19655}} (\bibinfo{year}{2024}).
\newblock


\bibitem[Zhang et~al\mbox{.}(2024b)]%
        {zhang2024rade}
\bibfield{author}{\bibinfo{person}{Baowen Zhang}, \bibinfo{person}{Chuan Fang}, \bibinfo{person}{Rakesh Shrestha}, \bibinfo{person}{Yixun Liang}, \bibinfo{person}{Xiaoxiao Long}, {and} \bibinfo{person}{Ping Tan}.} \bibinfo{year}{2024}\natexlab{b}.
\newblock \showarticletitle{RaDe-GS: Rasterizing Depth in Gaussian Splatting}.
\newblock \bibinfo{journal}{\emph{arXiv preprint arXiv:2406.01467}} (\bibinfo{year}{2024}).
\newblock


\bibitem[Zhang et~al\mbox{.}(2022)]%
        {zhang20223dilg}
\bibfield{author}{\bibinfo{person}{Biao Zhang}, \bibinfo{person}{Matthias Nie{\ss}ner}, {and} \bibinfo{person}{Peter Wonka}.} \bibinfo{year}{2022}\natexlab{}.
\newblock \showarticletitle{3dilg: Irregular latent grids for 3d generative modeling}.
\newblock \bibinfo{journal}{\emph{Advances in Neural Information Processing Systems}}  \bibinfo{volume}{35} (\bibinfo{year}{2022}), \bibinfo{pages}{21871--21885}.
\newblock


\bibitem[Zhang et~al\mbox{.}(2023)]%
        {zhang20233dshape2vecset}
\bibfield{author}{\bibinfo{person}{Biao Zhang}, \bibinfo{person}{Jiapeng Tang}, \bibinfo{person}{Matthias Niessner}, {and} \bibinfo{person}{Peter Wonka}.} \bibinfo{year}{2023}\natexlab{}.
\newblock \showarticletitle{3dshape2vecset: A 3d shape representation for neural fields and generative diffusion models}.
\newblock \bibinfo{journal}{\emph{ACM Transactions on Graphics (TOG)}} \bibinfo{volume}{42}, \bibinfo{number}{4} (\bibinfo{year}{2023}), \bibinfo{pages}{1--16}.
\newblock


\bibitem[Zhang et~al\mbox{.}(2024c)]%
        {zhang2024clay}
\bibfield{author}{\bibinfo{person}{Longwen Zhang}, \bibinfo{person}{Ziyu Wang}, \bibinfo{person}{Qixuan Zhang}, \bibinfo{person}{Qiwei Qiu}, \bibinfo{person}{Anqi Pang}, \bibinfo{person}{Haoran Jiang}, \bibinfo{person}{Wei Yang}, \bibinfo{person}{Lan Xu}, {and} \bibinfo{person}{Jingyi Yu}.} \bibinfo{year}{2024}\natexlab{c}.
\newblock \showarticletitle{CLAY: A Controllable Large-scale Generative Model for Creating High-quality 3D Assets}.
\newblock \bibinfo{journal}{\emph{ACM Transactions on Graphics (TOG)}} \bibinfo{volume}{43}, \bibinfo{number}{4} (\bibinfo{year}{2024}), \bibinfo{pages}{1--20}.
\newblock


\bibitem[Zhao et~al\mbox{.}(2024)]%
        {zhao2024michelangelo}
\bibfield{author}{\bibinfo{person}{Zibo Zhao}, \bibinfo{person}{Wen Liu}, \bibinfo{person}{Xin Chen}, \bibinfo{person}{Xianfang Zeng}, \bibinfo{person}{Rui Wang}, \bibinfo{person}{Pei Cheng}, \bibinfo{person}{Bin Fu}, \bibinfo{person}{Tao Chen}, \bibinfo{person}{Gang Yu}, {and} \bibinfo{person}{Shenghua Gao}.} \bibinfo{year}{2024}\natexlab{}.
\newblock \showarticletitle{Michelangelo: Conditional 3d shape generation based on shape-image-text aligned latent representation}.
\newblock \bibinfo{journal}{\emph{Advances in Neural Information Processing Systems}}  \bibinfo{volume}{36} (\bibinfo{year}{2024}).
\newblock


\bibitem[Zhou et~al\mbox{.}(2024)]%
        {zhou2024diffgs}
\bibfield{author}{\bibinfo{person}{Junsheng Zhou}, \bibinfo{person}{Weiqi Zhang}, {and} \bibinfo{person}{Yu-Shen Liu}.} \bibinfo{year}{2024}\natexlab{}.
\newblock \showarticletitle{Diffgs: Functional gaussian splatting diffusion}.
\newblock \bibinfo{journal}{\emph{arXiv preprint arXiv:2410.19657}} (\bibinfo{year}{2024}).
\newblock


\bibitem[Zou et~al\mbox{.}(2024)]%
        {zou2024triplane}
\bibfield{author}{\bibinfo{person}{Zi-Xin Zou}, \bibinfo{person}{Zhipeng Yu}, \bibinfo{person}{Yuan-Chen Guo}, \bibinfo{person}{Yangguang Li}, \bibinfo{person}{Ding Liang}, \bibinfo{person}{Yan-Pei Cao}, {and} \bibinfo{person}{Song-Hai Zhang}.} \bibinfo{year}{2024}\natexlab{}.
\newblock \showarticletitle{Triplane meets gaussian splatting: Fast and generalizable single-view 3d reconstruction with transformers}. In \bibinfo{booktitle}{\emph{Proceedings of the IEEE/CVF Conference on Computer Vision and Pattern Recognition}}. \bibinfo{pages}{10324--10335}.
\newblock


\bibitem[Zuo et~al\mbox{.}(2024)]%
        {zuo2024sparse3d}
\bibfield{author}{\bibinfo{person}{Qi Zuo}, \bibinfo{person}{Xiaodong Gu}, \bibinfo{person}{Yuan Dong}, \bibinfo{person}{Zhengyi Zhao}, \bibinfo{person}{Weihao Yuan}, \bibinfo{person}{Lingteng Qiu}, \bibinfo{person}{Liefeng Bo}, {and} \bibinfo{person}{Zilong Dong}.} \bibinfo{year}{2024}\natexlab{}.
\newblock \showarticletitle{High-Fidelity 3D Textured Shapes Generation by Sparse Encoding and Adversarial Decoding}. In \bibinfo{booktitle}{\emph{European Conference on Computer Vision}}.
\newblock


\end{thebibliography}

\newpage
\appendix
\begin{figure*}
  \includegraphics[width=\textwidth]{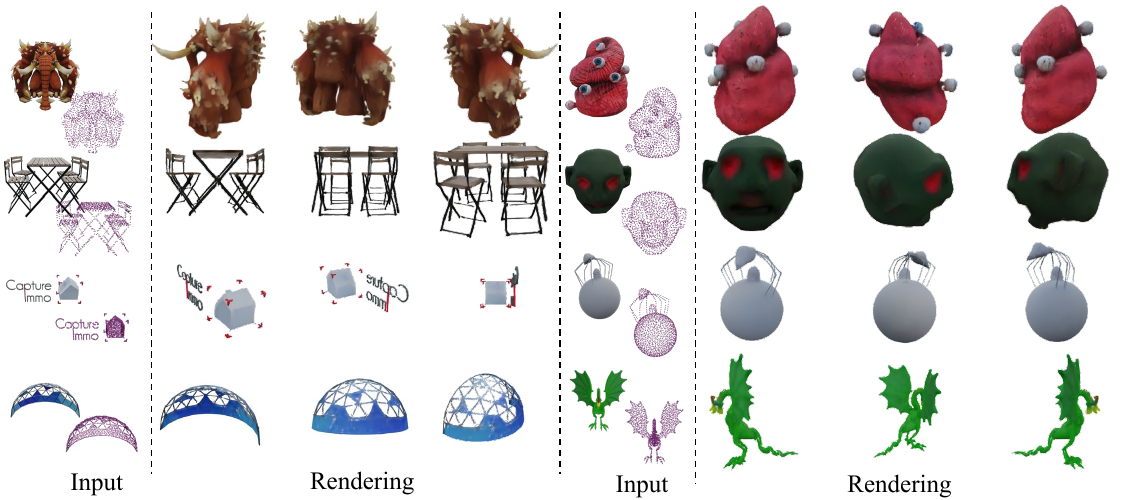}
  \vspace{-1.5\intextsep}
  \caption{The Reconstruction results of Anchor-GS VAE}
  \label{fig:vae}
\end{figure*}

\begin{figure*}
  \includegraphics[width=0.95\textwidth]{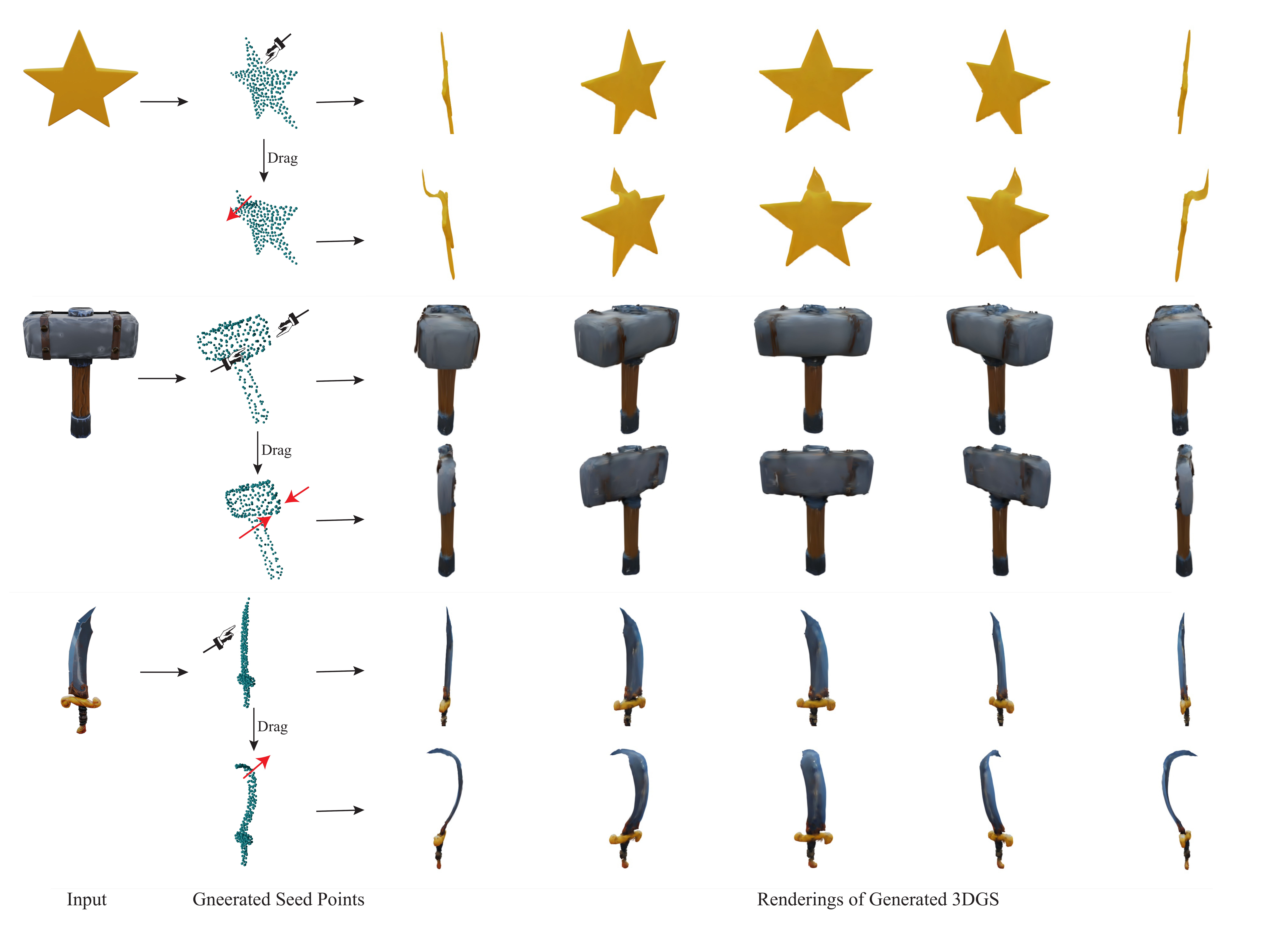}
  \vspace{-2\intextsep}
  \caption{Results of drag-based editing: By performing a limited number of drag edits on the seed points, we achieve finely controlled, edited results.}
  \label{fig:edit}
\end{figure*}

\begin{figure*}
  \includegraphics[width=0.95\textwidth]{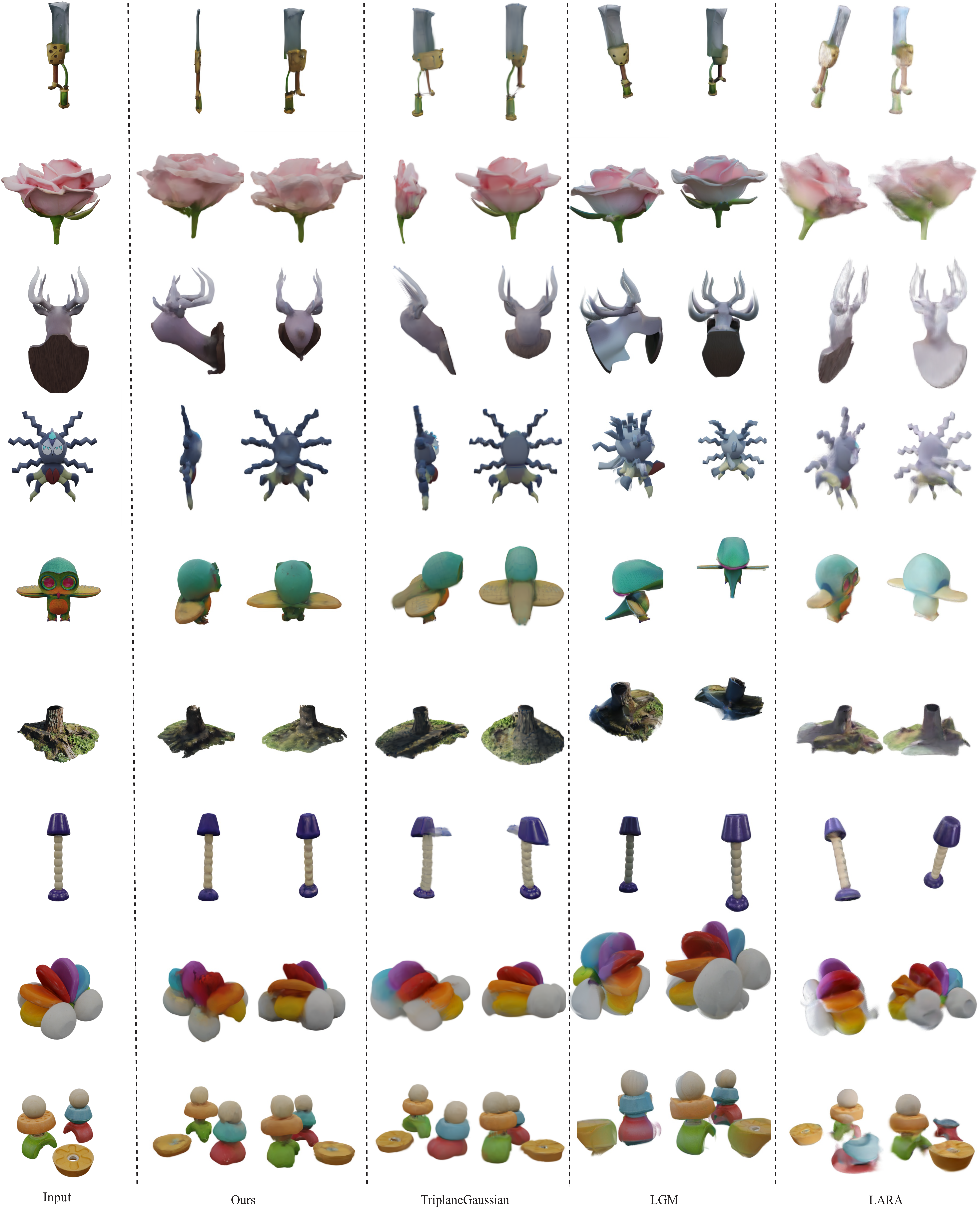}
  \caption{Comparison of single-image to 3DGS generation: Our method achieves more multi-view geometrically consistent results.}
  \label{fig:image-3d}
\end{figure*}

\end{document}